\documentclass[journal]{IEEEtran}

\usepackage{times}
\usepackage{epsfig}
\usepackage{graphicx}
\usepackage{amsmath}
\usepackage{amssymb}
\usepackage{mathrsfs}
\usepackage{multirow}
\usepackage{blkarray}
\usepackage{float}
\usepackage{array}
\usepackage{epstopdf}
\usepackage{subcaption}
\usepackage[T1]{fontenc}
\usepackage{algorithm}
\usepackage{algpseudocode}

\begin{document}

\title{{Text Extraction From Texture Images Using Masked Signal Decomposition}}

\author{Shervin~Minaee,~\IEEEmembership{Student Member,~IEEE,}
        and~Yao~Wang,~\IEEEmembership{Fellow,~IEEE}
}

\maketitle

\begin{abstract}
Text extraction is an important problem in image processing with applications from optical character recognition to autonomous driving.
Most of the traditional text segmentation algorithms consider separating text from a simple background (which usually has a different color from texts).
In this work we consider separating texts from a textured background, that has similar color to texts.
We look at this problem from a signal decomposition perspective, and consider a more realistic scenario where signal components are overlaid on top of each other (instead of adding together).
When the signals are overlaid, to separate signal components, we need to find a binary mask which shows the support of each component.
Because directly solving the binary mask is intractable, we relax this problem to the approximated continuous problem, and solve it by alternating optimization method. 
We show that the proposed algorithm achieves significantly better results than other recent works on several challenging images.
\end{abstract}


\IEEEpeerreviewmaketitle

\section{Introduction}
Text extraction is an important problem with many applications in image processing and computer vision, such as optical character recognition, license plate detection, road sign detection in autonomous driving. 
Text extraction could be very challenging when the background has complicated texture and has overlapping color distributions as the text.
Text extraction is usually accomplished in two steps: text region detection which detects the regions where text is high likely to be present \cite{detect}, and text segmentation to find a binary mask which shows the location of text.
Our main focus in this work would be on the second step, which is to derive a binary mask for text segmentation in a detected region containing texts.

Different algorithms have been proposed in the past for text segmentation from images.
In \cite{djvu}, Haffner et al proposed an algorithm for text image segmentation and document compression using a hierarchical clustering approach.
In \cite{spec}, an algorithm is proposed for segmentation  of  texts and graphics from screen content images and coding.   
In \cite{seg1}, Kumar proposes an algorithm for extracting texts from document images using matched wavelet filters.
They used a clustering-based technique to estimate globally matched wavelet filters using a collection of groundtruth images.
In \cite{seg2}, Saha proposed a text segmentation algorithm using Hough transform.
Various algorithms based on sparse decomposition has also been proposed for text extraction, where the text extraction is achieved by assuming proper prior on the text component \cite{sp_text}, \cite{myTV}. 
There are also many other works based on histogram analysis, maximally stable extremal region (MSER), and appearance \cite{text1}-\cite{text5}.
The reader is referred to \cite{text6} for a good survey of text recognition.

In this work we look at this problem from a signal decomposition perspective.
We assume that the foreground texts $x_2$, and background $x_1$ are combined to create an image $x$.
But instead of assuming an additive model  ($x= x_1+x_2$) as in \cite{sp_text}-\cite{myTV}, we consider an overlaying model, which is a  more truthful characterization of images with overlaid texts.
Specifically, we assume any pixel value in the image comes either from the background component, or from foreground text.
We can formulate this as:
\begin{equation}
x= (1-w) \circ x_1+ w \circ x_2  \ \ \ \ \text{s.t.}  \ \ w \in \{0,1\}^n
\end{equation}
where $\circ$ denotes the element-wise product, and $w$ is the binary mask indicating foreground pixel locations. 

If we have some prior knowledge about each component, we can solve the decomposition problem as the following optimization problem:
\begin{equation}
\begin{aligned}
& \hspace{-0.2cm}\underset{w, x_1, x_2}{\text{min}}
 \phi_1(x_1)+ \lambda_1 \phi_2(x_2)+ \lambda_2 \phi_3(w) \\
& \ \text{s.t.}
\ \  w \in \{0,1\}^n  , x= (1-w) \circ x_1+ w \circ x_2
\end{aligned}
\end{equation}
where $\phi_i$ are some cost functions that should be minimized based on our prior knowledge about $x_1$, $x_2$, and $w$.
After solving this optimization problem, $w$ will give us the location of texts.
We will discuss further about the problem formulation and solution in the next section.

Figure 2 compares the segmentation results using hierarchical k-means approach, sparse decomposition, and the proposed algorithm in this work for a sample image.
As we can see, each of the previous approaches have their own difficulties, for example clustering based scheme would have difficulty separating text from background in the case where text has a similar color to background. 
And sparse decomposition based model, misses some part of the text, while detecting some part of the background as text. 
On the other hand, the proposed algorithm performs significantly better.
This approach can also be used for segmentation of medical images \cite{MD1}, and also texture extraction from biometrics \cite{my1}-\cite{my2}.

\begin{figure}
        \centering
        \hspace{-0.2cm}
        \begin{subfigure}[b]{0.22\textwidth}
                \includegraphics[width=\textwidth]{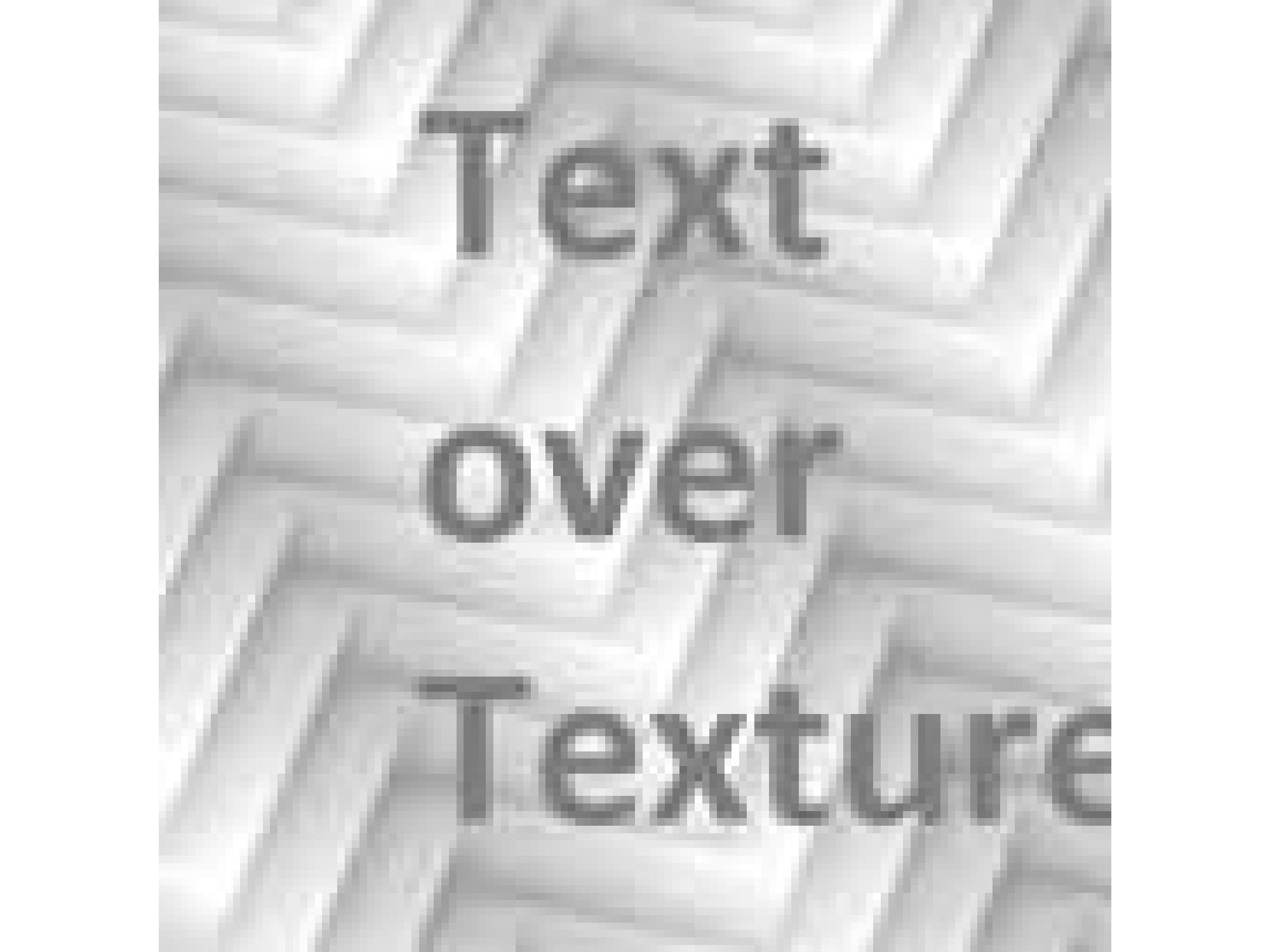}
                \vspace{-0.5cm}
            \hspace{-3.5cm} 
        \end{subfigure}%
        ~ 
        \begin{subfigure}[b]{0.22\textwidth}
                \includegraphics[width=\textwidth]{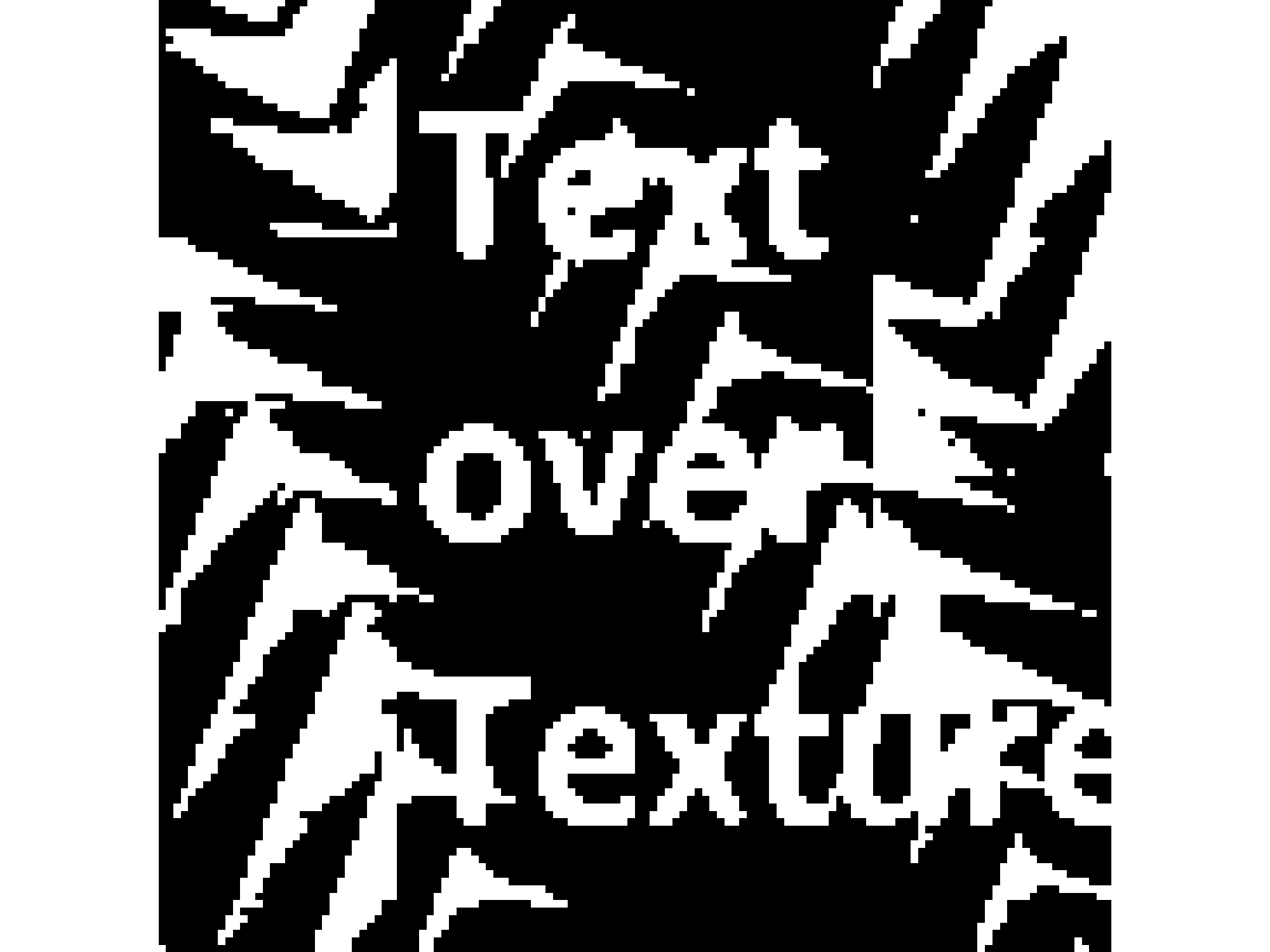}
                 \vspace{-0.45cm}
              \hspace{-4.8cm}
        \end{subfigure}
         \\[1ex]
                 \centering
        \vspace{0.1cm}
        \hspace{-1.4cm}

        \begin{subfigure}[b]{0.22\textwidth}
                \includegraphics[width=\textwidth]{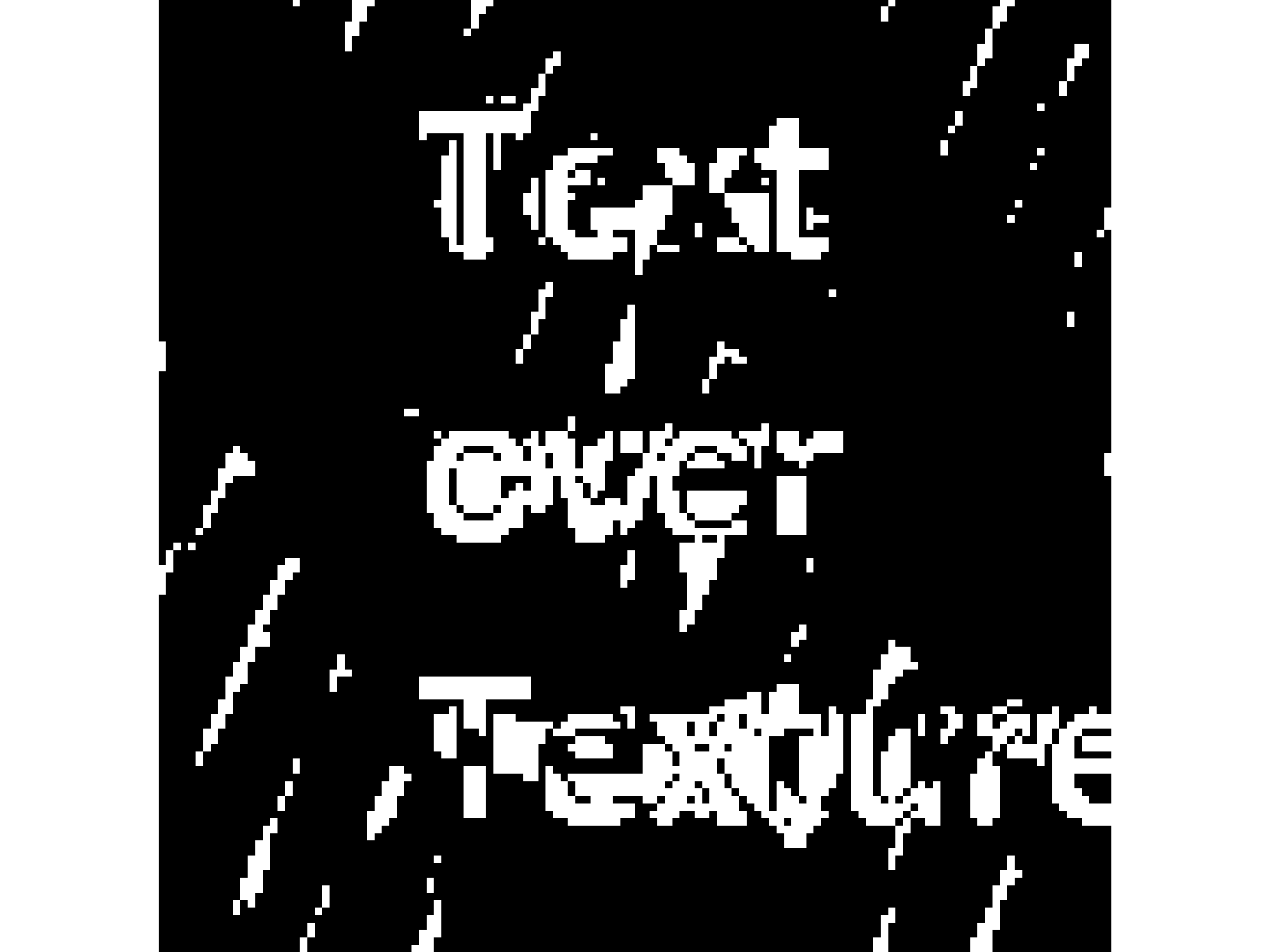}
                \vspace{-0.5cm}
            \hspace{-3cm} 
        \end{subfigure}%
        ~ 
        \begin{subfigure}[b]{0.22\textwidth}
                \includegraphics[width=\textwidth]{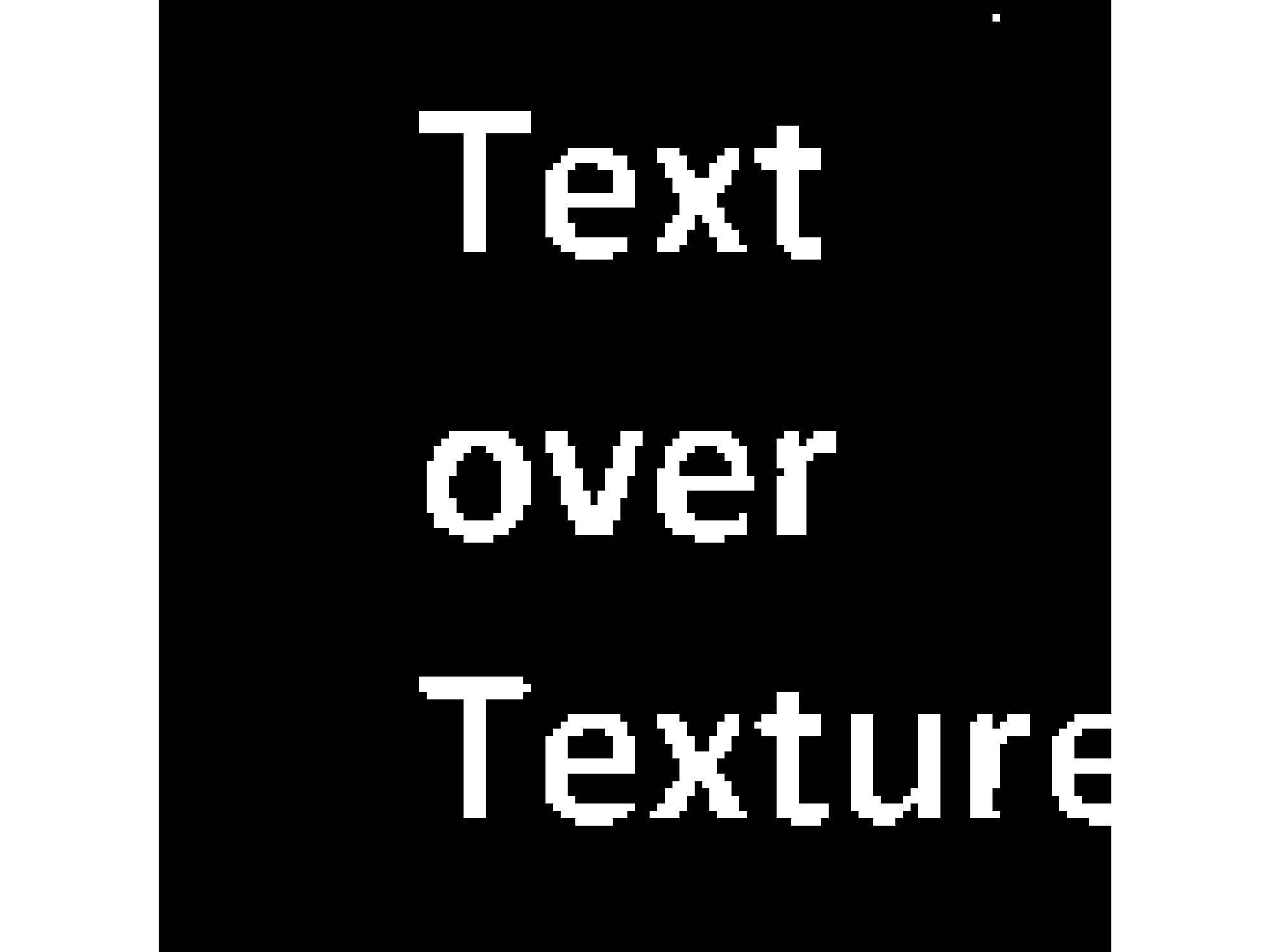}
                 \vspace{-0.45cm}
              \hspace{-4.8cm}
        \end{subfigure}
        \vspace{0.45cm}
         \\[1ex]
        \caption{ The top-right, bottom-left and bottom-right images denote the segmentation map by hierarchical clustering \cite{djvu}, sparse decomposition \cite{myTV}, and the proposed algorithm.}
\end{figure}

The rest of this paper is structured as follows: Section II presents the problem formulation, and the iterative algorithm to solve the optimization problem.
The experimental results and performance analysis are provided in Section III, and the paper is concluded in Section IV.

\section{Problem Formulation}
As discussed earlier, we consider text as an overlaid component on the background image. Therefore we can denote an image as the masked summation of the background and text:
\begin{equation}
x= (1-w) \circ x_1+ w \circ  x_2,  \ \ \ \ w \in \{0,1\}^n
\end{equation}

To further simplify the problem, we assume that each component has a sparse representation using some properly designed dictionary or forming a subspace.
Therefore we can write this problem as:
\begin{equation}
x= (1-w) \circ {(P_1\alpha_1)}+ w \circ  (P_2\alpha_2)  \ \ \  s.t.\ \  \ \ \| \alpha_i \|_0 \leq k_i
\end{equation}
where $P_i$ is n$\times \text{m}_i$ matrix, where each column denotes one of the basis functions from the corresponding subspace/dictionary.
We would like to note that here we assume that both subspaces are known, in the applications where the choice of subspaces/dictionaries is not clear, we can use dictionary/subspace learning algorithms to learn them \cite{dic1}-\cite{dic6}.

Note that, the model in (4) can be rewritten as:
\begin{equation}
x= (I-W) {P_1\alpha_1}+ W P_2\alpha_2 \ \ \  s.t.\ \  \ \ \| \alpha_i \|_0 \leq k_i
\end{equation}
where $W$ is a diagonal matrix with the vector $w$ on its main diagonal (i.e. $W= \text{diag}(w)$). If a diagonal element is 1, the corresponding pixel belongs to the foreground, and otherwise to the background.

The decomposition problem in Eq. (4) is a highly ill-posed problem. 
Therefore we need to impose some prior on each component, and also on $w$ to be able to perform this decomposition.
We assume that each component has a sparse representation with respect to its own subspace, but not with respect to the other one.
We also assume that the second component is sparse, which is very desirable for text. 
To promote sparsity of the second component, we add the $\ell_0$ norm of $w$ to the cost function (note that $w$ corresponds to the support of the second component).

We can incorporate all these priors in an optimization problem as shown below:

\begin{small}
\begin{equation}
\begin{aligned}
& \hspace{-0.2cm}\underset{w, \alpha_1, \alpha_2}{\text{min}}
 \  \ \frac{1}{2} \| f- (1-w)\circ  P_1\alpha_1- w \circ  P_2\alpha_2  \|_2^2+ \lambda \| w \|_0 \\
& \ \text{s.t.}
\ \ \ \ \ \ w \in \{0,1\}^n  , \ \| \alpha_1 \|_0 \leq k_1 , \ \| \alpha_2 \|_0 \leq k_2
\end{aligned}
\end{equation}
\end{small}

This problem is a combinatorial problem, and is not tractable, both because of the  $\| w \|_0$ term in the cost function and also the binary nature of $w $. We relax these conditions to be able to solve this problem in an alternating optimization approach. 
We replace the $\| w \|_0$ in the cost function with $\| w \|_1$, and also relax the $w \in \{0,1\}^n$ condition to $w \in [0,1]^n$.
Then we will get the following optimization problem:

\begin{small}
\begin{equation}
\begin{aligned}
& \hspace{-0.2cm}\underset{w, \alpha_1, \alpha_2}{\text{min}}
 \  \ \frac{1}{2} \| f- (1-w)\circ  P_1\alpha_1- w \circ  P_2\alpha_2  \|_2^2+ \lambda \| w \|_1 \\
& \ \text{s.t.}
\ \ \ \ \ \ w \in [0,1]^n  , \ \| \alpha_1 \|_0 \leq k_1 , \ \| \alpha_2 \|_0 \leq k_2
\end{aligned}
\end{equation}
\end{small}

This problem can be solved with different approaches, such as majorization minimization, alternating direction method, and random sampling approach \cite{lag1}-\cite{ransac2}.
Here we present an algorithm based on alternating direction method, which simply solves this problem by updating each variable at a time, and setting the gradient of cost function ($L$) w.r.t. each variable to zero.
The optimization steps with respect to $\alpha_1$ and $\alpha_2$ are symmetric, therefore  we only show the solution for $\alpha_2$.
We first ignore the constraint $ \| \alpha_2 \|_0 \leq k_2$ and solve the unconstrained problem, and then keep the largest $k_2$ components of $\alpha_2$ to satisfy the constraint. 
\begin{equation*}
\begin{aligned}
  \alpha_2=    \underset{ \alpha_2}{\text{\ \ argmin}} 
 \  \| f- (I-W) P_1\alpha_1- W  P_2\alpha_2  \|_2^2 \Rightarrow \hspace{1.2cm}\\
  \nabla_{\alpha_2}L=0 \Rightarrow P_2^tW^t \big( WP_2\alpha_2+(I-W)P_1\alpha_1-f \big)=0 \hspace{0.5cm}  \\ 
 \Rightarrow \alpha_2= (P_2^tW^tWP_2)^{-1} P_2^tW^t (f-(I-W)P_1\alpha_1) \hspace{1.18cm} 
\end{aligned}
\end{equation*}
Then by keeping the $k_2$ largest components of the above solution, we will derive the $\alpha_2$ update as $\alpha_2^*= \Pi_{top}(\alpha_2; k_2)$.
We also provide the update step of $w$ here.
By using the equality $\text{diag}(w) P_2\alpha_2= \text{diag}(P_2\alpha_2) w $, we can re-write the optimization w.r.t. $w$ as below:
\begin{equation*}
\begin{aligned}
& \underset{w}{\text{min}}
 \  \ \frac{1}{2} \| f- \text{diag}(P_1\alpha_1)(1-w) - \text{diag}(P_2\alpha_2) w  \|_2^2+ \lambda \| w \|_1   \\
& \ \text{s.t.}
\ \ \  w \in [0,1]^n  
\end{aligned}
\end{equation*}
First note that, we can simplify the above optimization problem as below:
\begin{equation}
\begin{aligned}
& \underset{w}{\text{min}}
 \  \ \frac{1}{2} \| h - C w  \|_2^2+ \lambda \| w \|_1   \\
& \ \text{s.t.}
\ \ \  w \in [0,1]^n  
\end{aligned}
\end{equation}
where $C= \text{diag}(P_2\alpha_2)-\text{diag}(P_1\alpha_1)= \text{diag}(P_2\alpha_2-P_1\alpha_1)$, and $h= f- P_1\alpha_1$.
We can first ignore the constraint, and then project the optimal solution of the cost function onto the feasible set ($w \in [0,1]^n$).
Since $C$ is a diagonal matrix, the cost function can be decoupled in the components of $w$ as:
\begin{equation}
\begin{aligned}
& \underset{w}{\text{min}}
 \  \ g(w)= \frac{1}{2} \| h - C w  \|_2^2+ \lambda \| w \|_1  \\
 & = \sum_{i=1}^n g_i(w_i)= \sum_{i=1}^n \frac{1}{2} |h_i- C_{ii}w_i|^2+ \lambda |w_i|_1
\end{aligned}
\end{equation}
where $C_{ii}$ denotes the i-th diagonal element of $C$ (which is the i-th diagonal element of the vector $P_1\alpha_1-   P_2\alpha_2$).
Now this problem can be easily solved with soft-thresholding as \cite{soft}:
\begin{equation}
\begin{aligned}
 \ w_i= \text{soft}( h_i/C_{ii}, \lambda/C^2_{ii})
\end{aligned} 
\end{equation}
where $\text{soft}(x,\lambda)$ denotes the soft-thresholding operator  applied element-wise and defined:
\begin{small}
\begin{gather*}
\text{Soft}(x,\lambda)= \text{sign}(x) \ \text{max}(|x|-\lambda,0)
\end{gather*}
\end{small}
Now if we denote the projection operator on [0,1] by $\Pi_{[0,1]}(w)$, then the optimization solution of $w$ step would be:
\begin{equation}
\begin{aligned}
 \ w_i^*= \Pi_{[0,1]}\big( \text{soft}( h_i/C_{ii}, \lambda/C^2_{ii}) \big) 
\end{aligned} 
\end{equation}

The overall algorithm is summarized in Algorithm 1.

\begin{algorithm}
  \caption{pseudo-code for variable updates of problem (7)}\label{euclid}
  \begin{algorithmic}[1]
  \begin{small}
      \For{\texttt{$k$=1:$k_{max}$}} \vspace{0.06cm} 
        \State $\alpha_1^{k+1}= \Pi_{top-k_1}\big( (P_1^t\bar{W}^t\bar{W}P_1)^{-1} P_1^t\bar{W}^t (f-WP_2\alpha_2) \big)$ \vspace{0.15cm}
        \State $\alpha_2^{k+1}=  \Pi_{top-k_1}\big( (P_2^tW^tWP_2)^{-1} P_2^tW^t (f-\bar{W} P_1\alpha_1) \big) $ \vspace{0.15cm} \hspace{0.5cm}
        \State $ w= \Pi_{[0,1]}\big( \text{soft}( h_i/C_{ii}, \lambda/C^2_{ii}) \big)      $    \vspace{0.15cm}       
      \EndFor
      \\ binarize $w$ by comparing to a threshold.
        \end{small}
  \end{algorithmic}
\end{algorithm}
where $\bar{W}= I-W$.

\section{Experimental Results}
In this section we provide the experimental study on the application of the proposed algorithm for text extraction from several challenging images. 
These images are manually generated by adding text on top of a relatively complicated background. 
Our dataset contains more than 300 image blocks.

We first define the parameters of our model.
We apply our algorithm on blocks of 64x64. We first convert each block into a vector of 4096 dimension, and then use the proposed algorithm for background foreground separation.
For the first component we use 40 dimensional low-frequency DCT subspace \cite{dct}, and for the second component we use 10 dimensional Hadamard subspace \cite{hada}.  The sparsity of coefficients are chosen as $k_1=5$ and $k_2=5$.
The weight parameter for the sparsity term is chosen to be $\lambda= 10$, which is tuned by testing on a validation set.
The number of iterations for optimization algorithm is chosen to be $k_{max}= 10$, and $w$ is initialized with uniform random variable in [0,1].

We now discuss two possible ways to perform the binarization.
As mentioned earlier, the goal is to solve the binary optimization problem, but to make it tractable, we approximate the binary variables with a continuous variable in $[0,1]$, and binarize them after solving the optimization problem.
In the algorithm 1, we first solve the optimization problem in Eq. (7), for a continuous $w$, and then binarize $w$ at the very end.
An alternative approach is to binarize the variables $w$ after each update of $w$ in algorithm 1.
We tested both approaches for two of the test images, and provided the results in Figure 2. 
As we can see, usually doing the binarization at the very end works better.
One possible reason could be that with the second approach, the final solution is very sensitive to the initialization.

\begin{figure}
        \centering
        \vspace{-0.5cm}
        \begin{subfigure}[b]{0.19\textwidth}
                \includegraphics[width=\textwidth]{texture8_orig-eps-converted-to.pdf}
                \vspace{-0.1cm}
            \hspace{-3.5cm} 
        \end{subfigure}%
        ~ 
        \begin{subfigure}[b]{0.25\textwidth}
                \includegraphics[width=\textwidth]{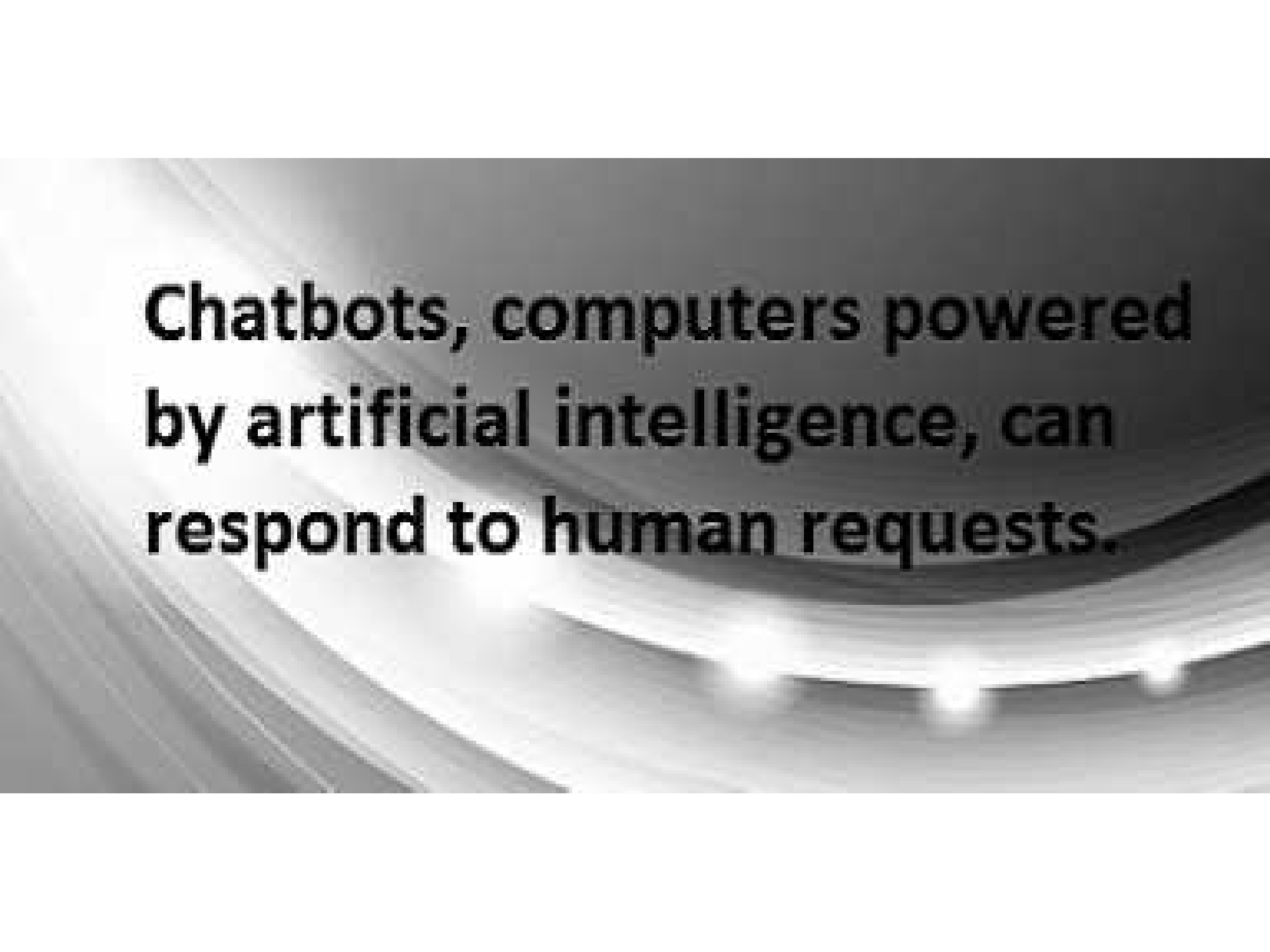}
                 \vspace{-0.45cm}
              \hspace{-4.8cm}
        \end{subfigure}
         \\[1ex]
                 \centering
        \vspace{-0.3cm}
        \hspace{-0.2cm}
        \begin{subfigure}[b]{0.19\textwidth}
                \includegraphics[width=\textwidth]{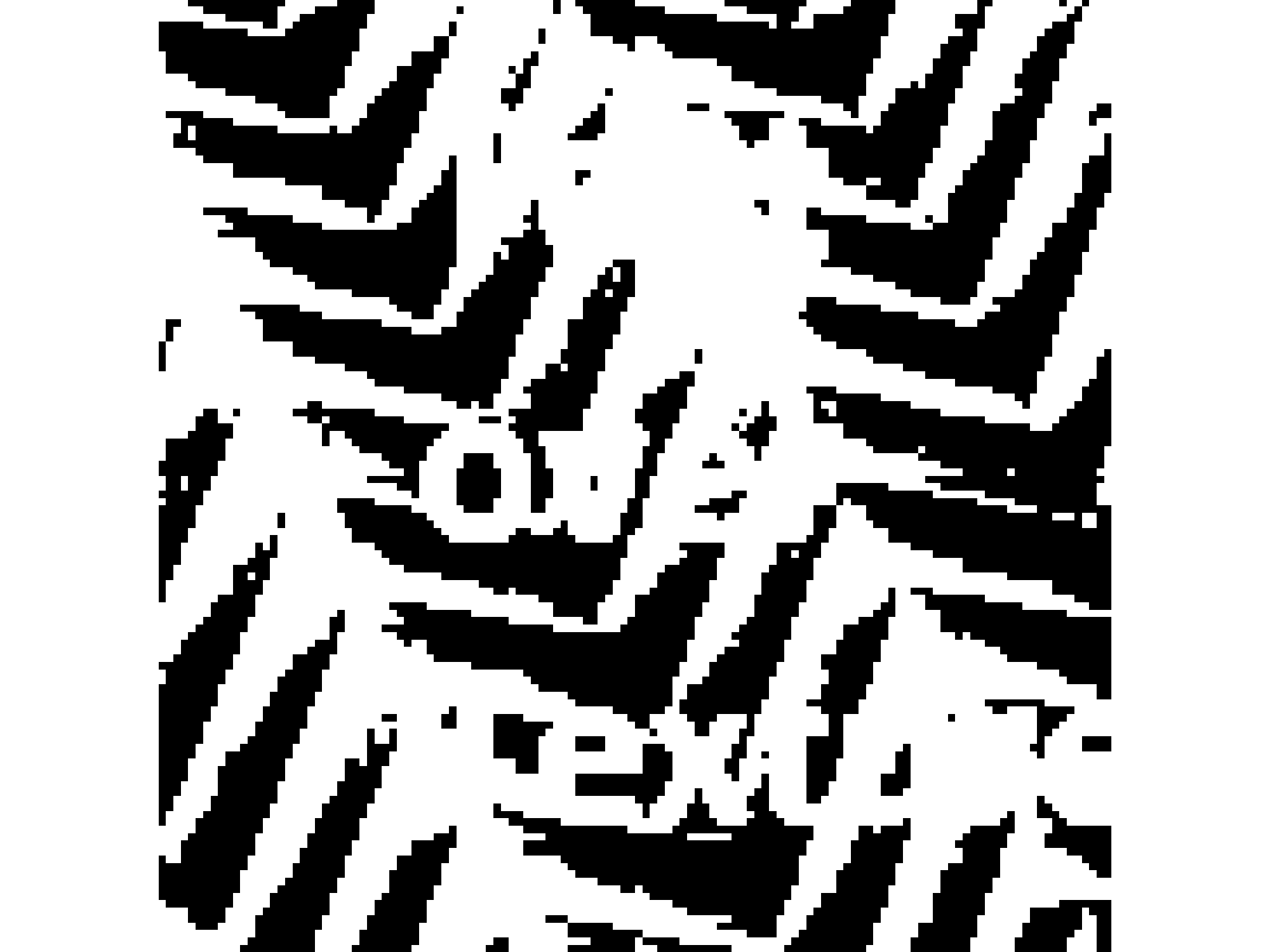}
                \vspace{-0.01cm}
            \hspace{-3cm} 
        \end{subfigure}%
        ~ 
        \begin{subfigure}[b]{0.25\textwidth}
                \includegraphics[width=\textwidth]{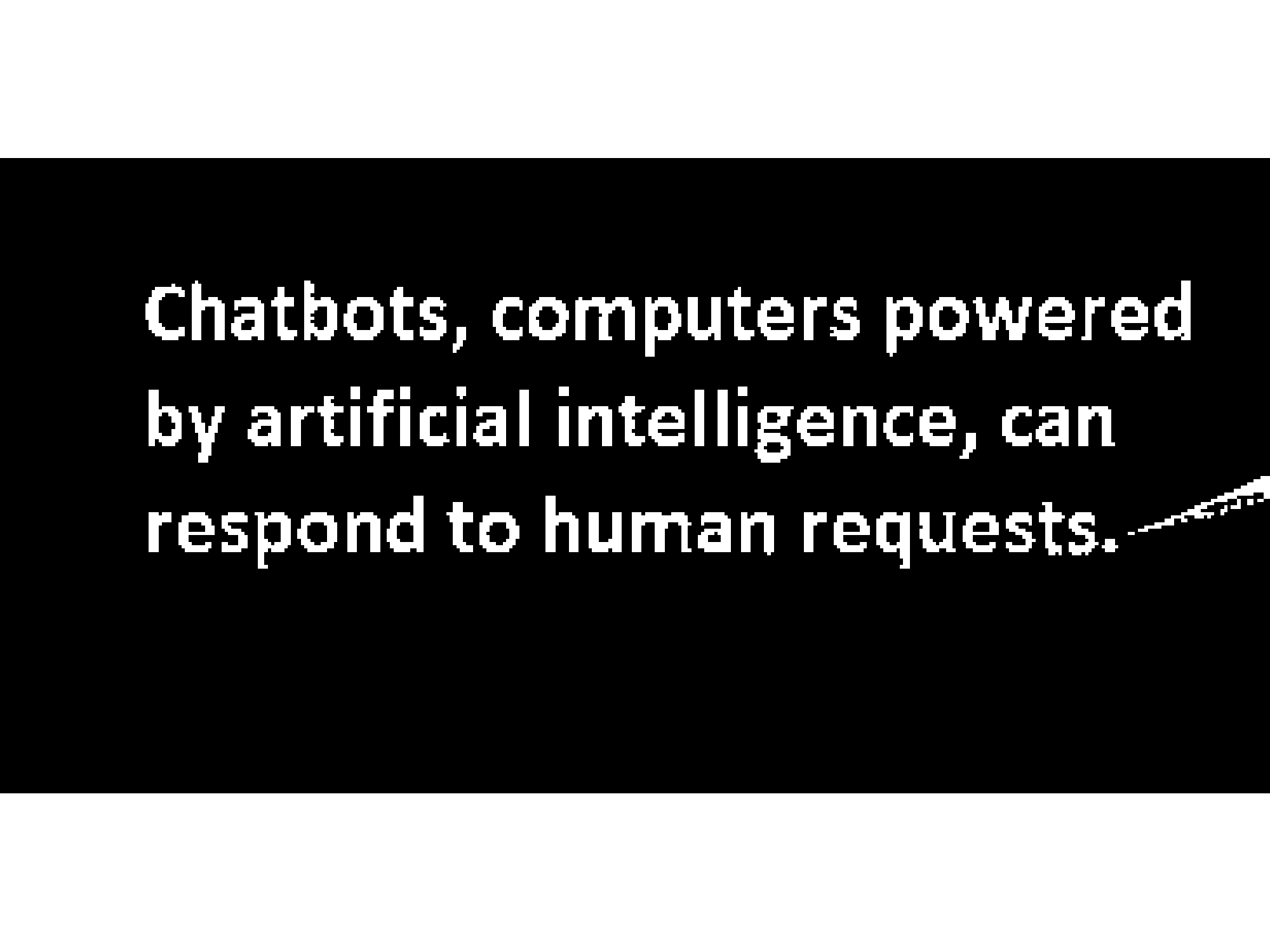}
                 \vspace{-0.45cm}
              \hspace{-4.8cm}
        \end{subfigure}
         \\[1ex]
                 \centering
        \vspace{-0.3cm}
        \hspace{-1.4cm}

        \begin{subfigure}[b]{0.19\textwidth}
       \vspace{-2cm}
                \includegraphics[width=\textwidth]{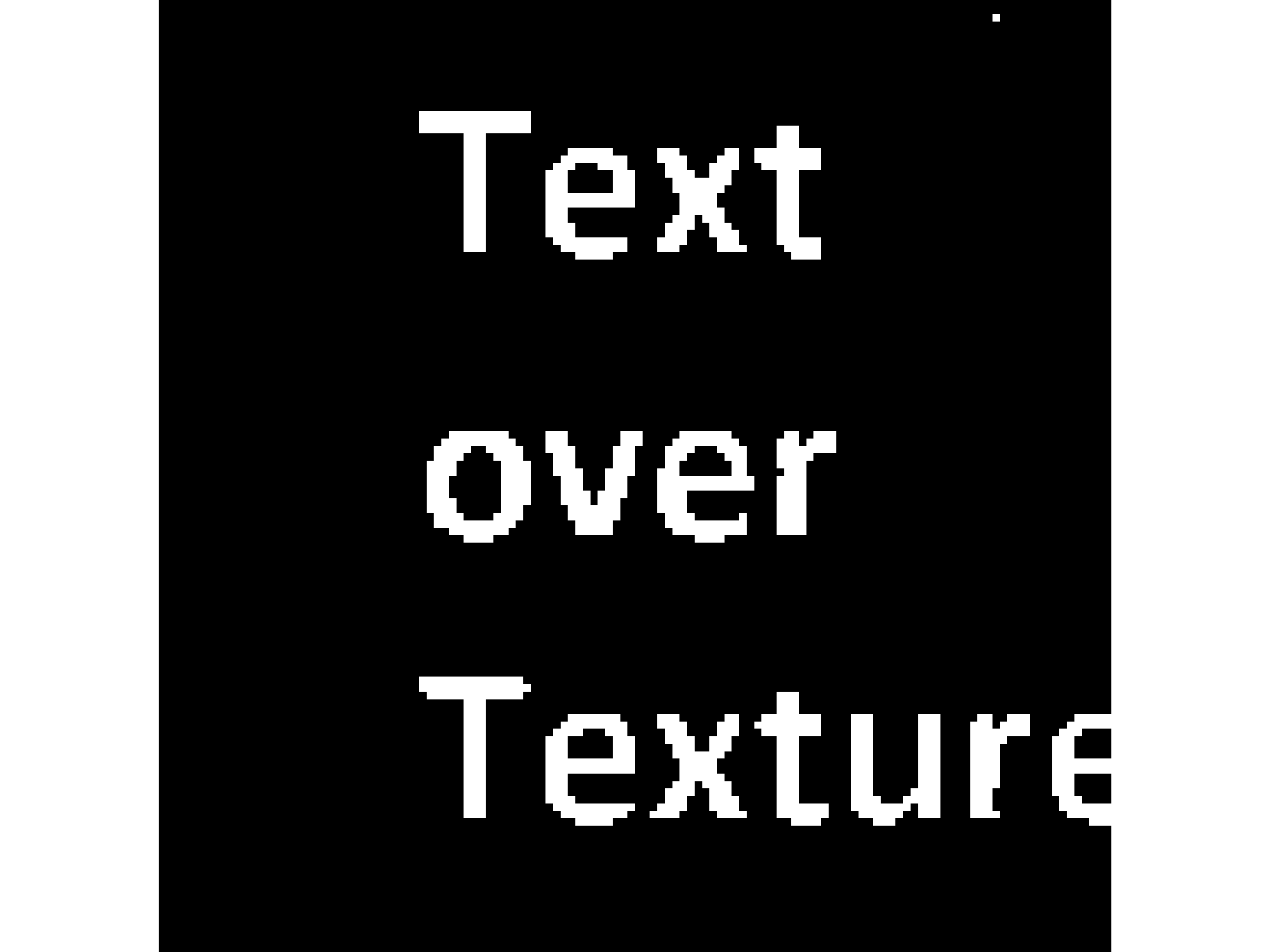}
                \vspace{-0.03cm}
            \hspace{-3cm} 
        \end{subfigure}%
        ~ 
        \begin{subfigure}[b]{0.25\textwidth}
       \vspace{-0.5cm}
                \includegraphics[width=\textwidth]{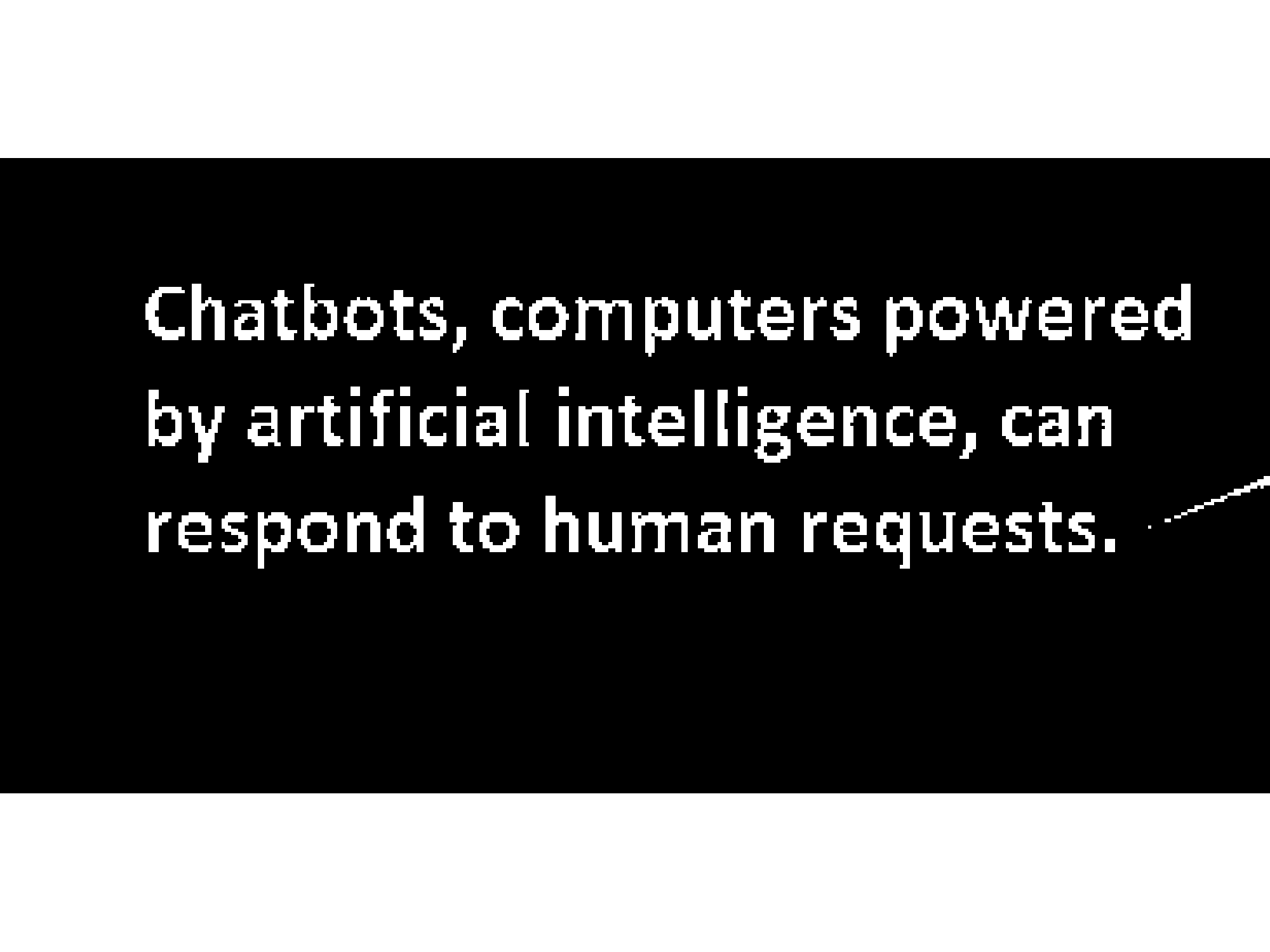}
                 \vspace{-0.45cm}
              \hspace{-4.8cm}
        \end{subfigure}   \\[1ex]
        \caption{ The images in the first to the third rows denote the original images, the foreground map using binarization in each iteration, and foreground map using binarization at the very end respectively.}
\end{figure}

We now provide the comparison of the proposed algorithm with prior approaches on text extraction.
We compare the proposed algorithm with three previous algorithms; hierarchical k-means clustering \cite{djvu}, shape primitive extraction, and sparsity based approach \cite{myTV}.
Figure 3 shows the comparison between the proposed algorithm performance compared with the previous approaches for 3 sample images from our dataset. As it can be seen, the proposed algorithm achieves more accurate result than previous methods.

\begin{figure*}
        \centering
        \vspace{-0.1cm}
        \begin{subfigure}[b]{0.22\textwidth}
                \includegraphics[width=\textwidth]{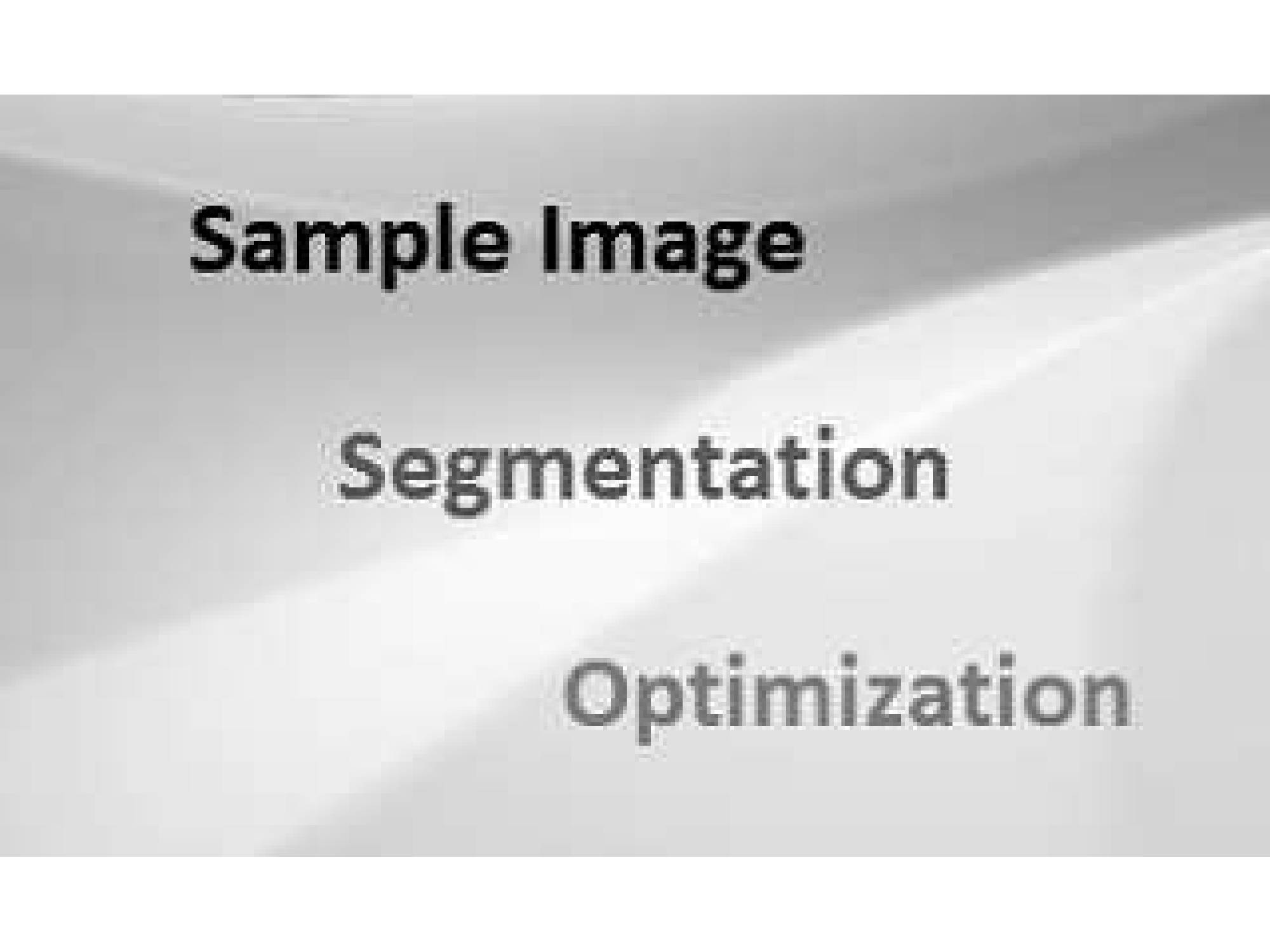}
                                \vspace{-0.3cm}
          \hspace{-1.5cm}    
        \end{subfigure}%
        ~ 
		\vspace{-0.02cm}        
        \begin{subfigure}[b]{0.18\textwidth}
                \includegraphics[width=\textwidth]{texture8_orig-eps-converted-to.pdf}
                \vspace{-0.04cm}
            \hspace{-6cm} 
        \end{subfigure}%
        \begin{subfigure}[b]{0.25\textwidth}
			~ 
            \vspace{-0.42cm}
                \includegraphics[width=\textwidth]{test2_orig-eps-converted-to.pdf}
                \vspace{-0.4cm}
            \hspace{-2cm} 
        \end{subfigure}%
        \begin{subfigure}[b]{0.30\textwidth}
                \includegraphics[width=\textwidth]{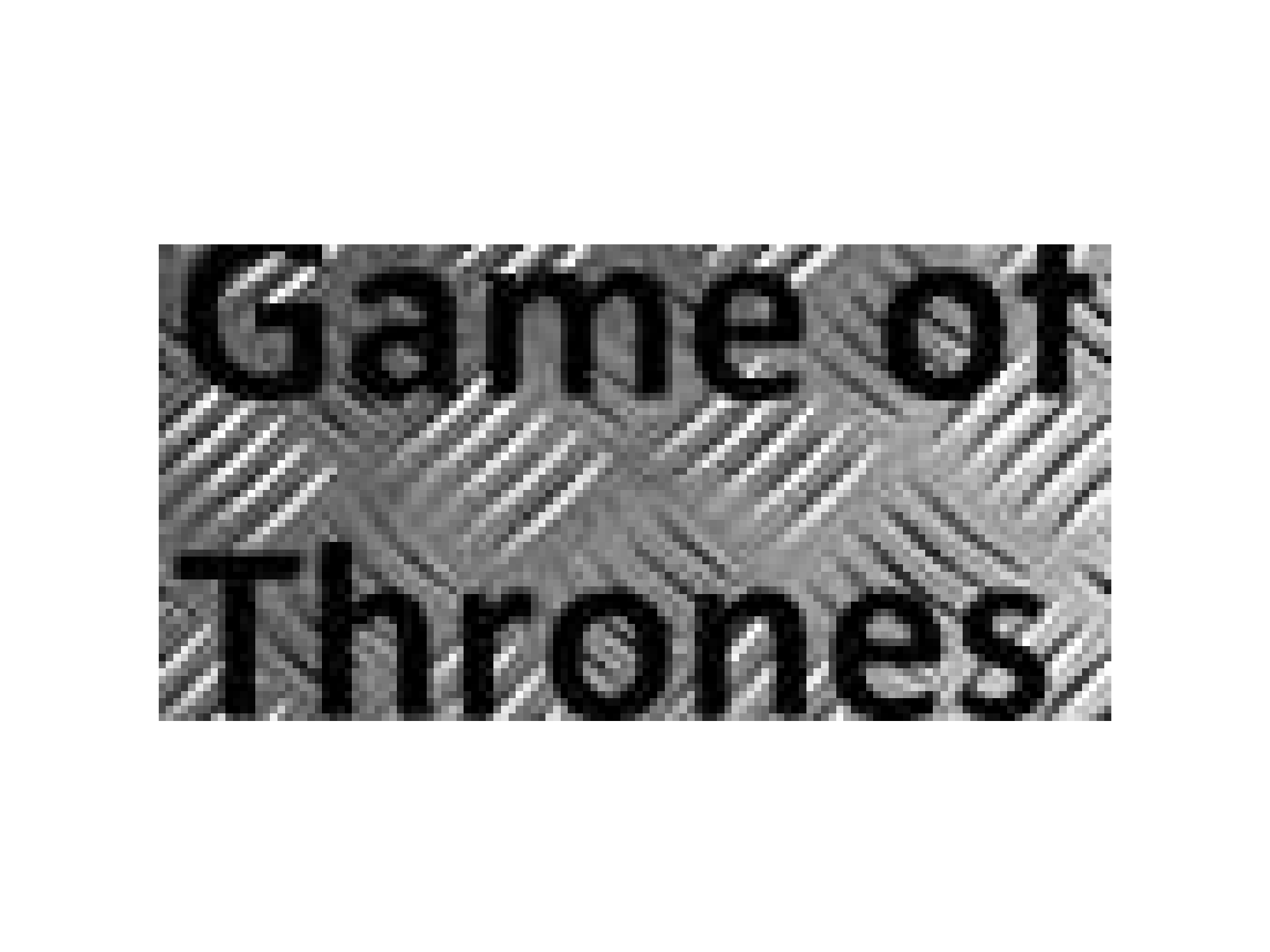}
                 \vspace{-0.61cm}
              \hspace{-4.8cm}
        \end{subfigure}
         \\[1ex]  \vspace{-0.9cm}
        \begin{subfigure}[b]{0.22\textwidth}
                \includegraphics[width=\textwidth]{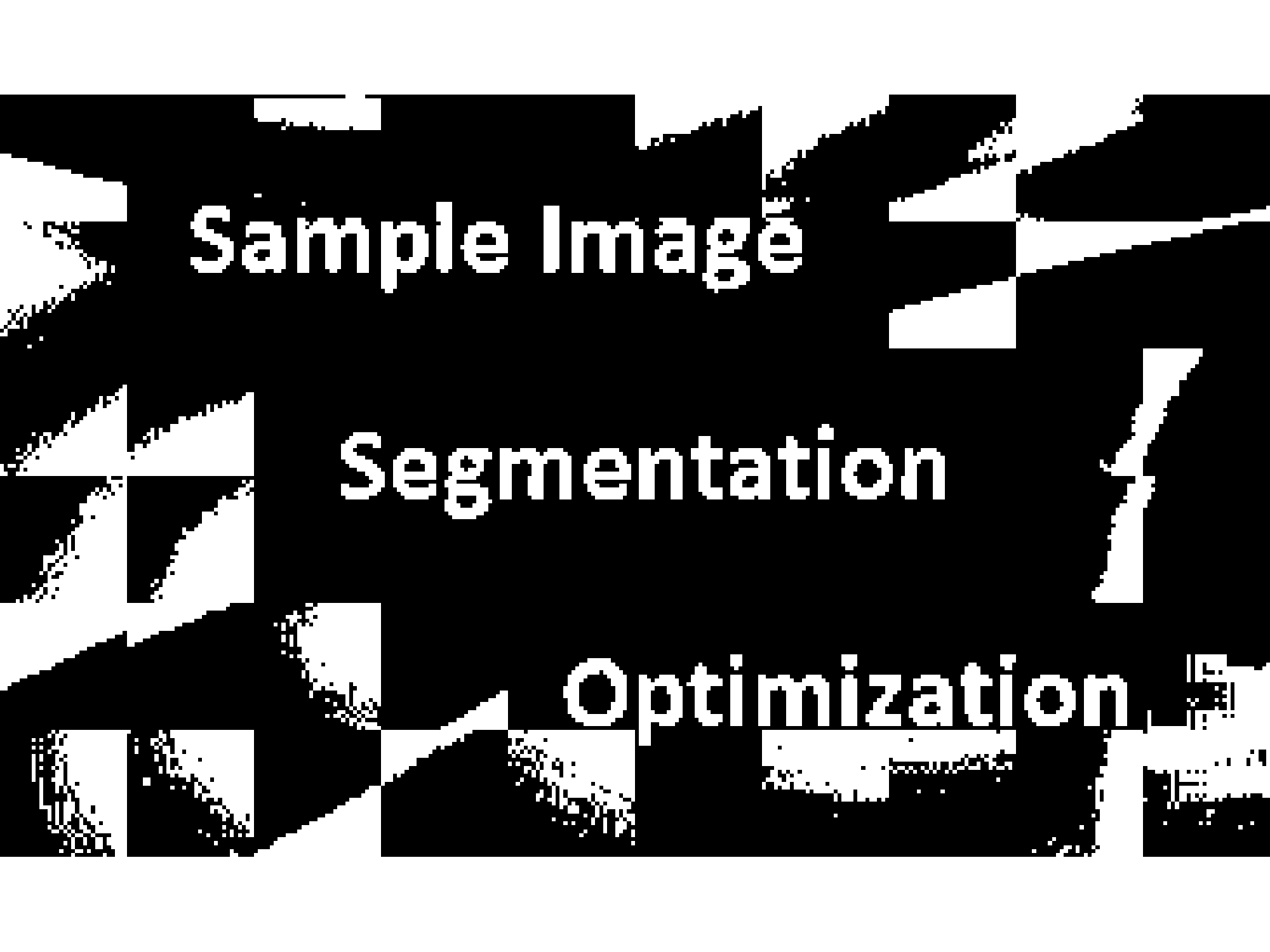}
                                \vspace{-0.32cm}
          \hspace{-1.5cm}    
        \end{subfigure}%
        ~ 
		\vspace{-0.02cm}        
        \begin{subfigure}[b]{0.18\textwidth}
                \includegraphics[width=\textwidth]{textuer8_DjVu-eps-converted-to.pdf}
                \vspace{-0.04cm}
            \hspace{-6cm} 
        \end{subfigure}%
        \begin{subfigure}[b]{0.25\textwidth}
			~ 
            \vspace{-0.42cm}
                \includegraphics[width=\textwidth]{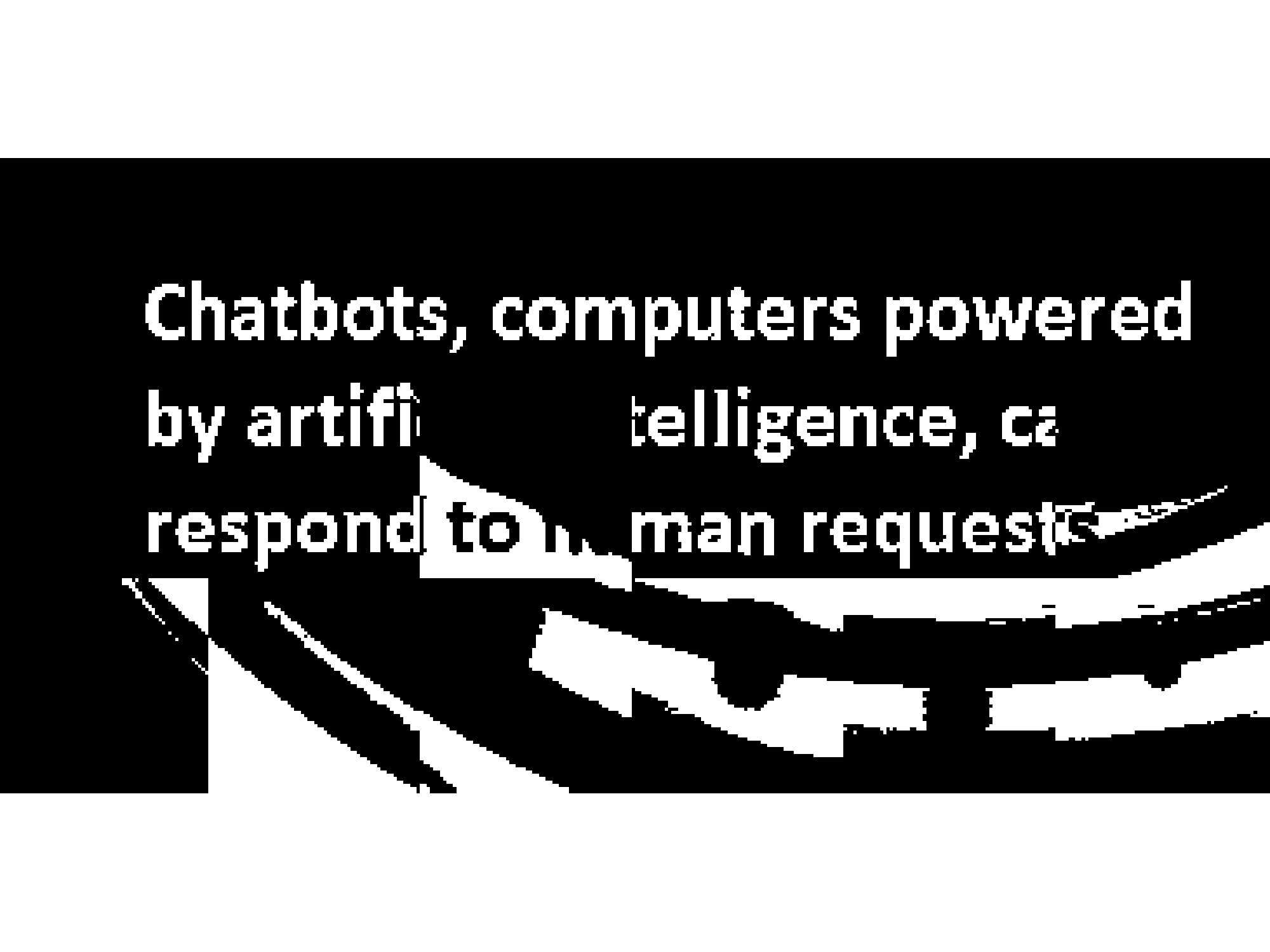}
                \vspace{-0.4cm}
            \hspace{-2cm} 
        \end{subfigure}%
        \begin{subfigure}[b]{0.30\textwidth}
                \includegraphics[width=\textwidth]{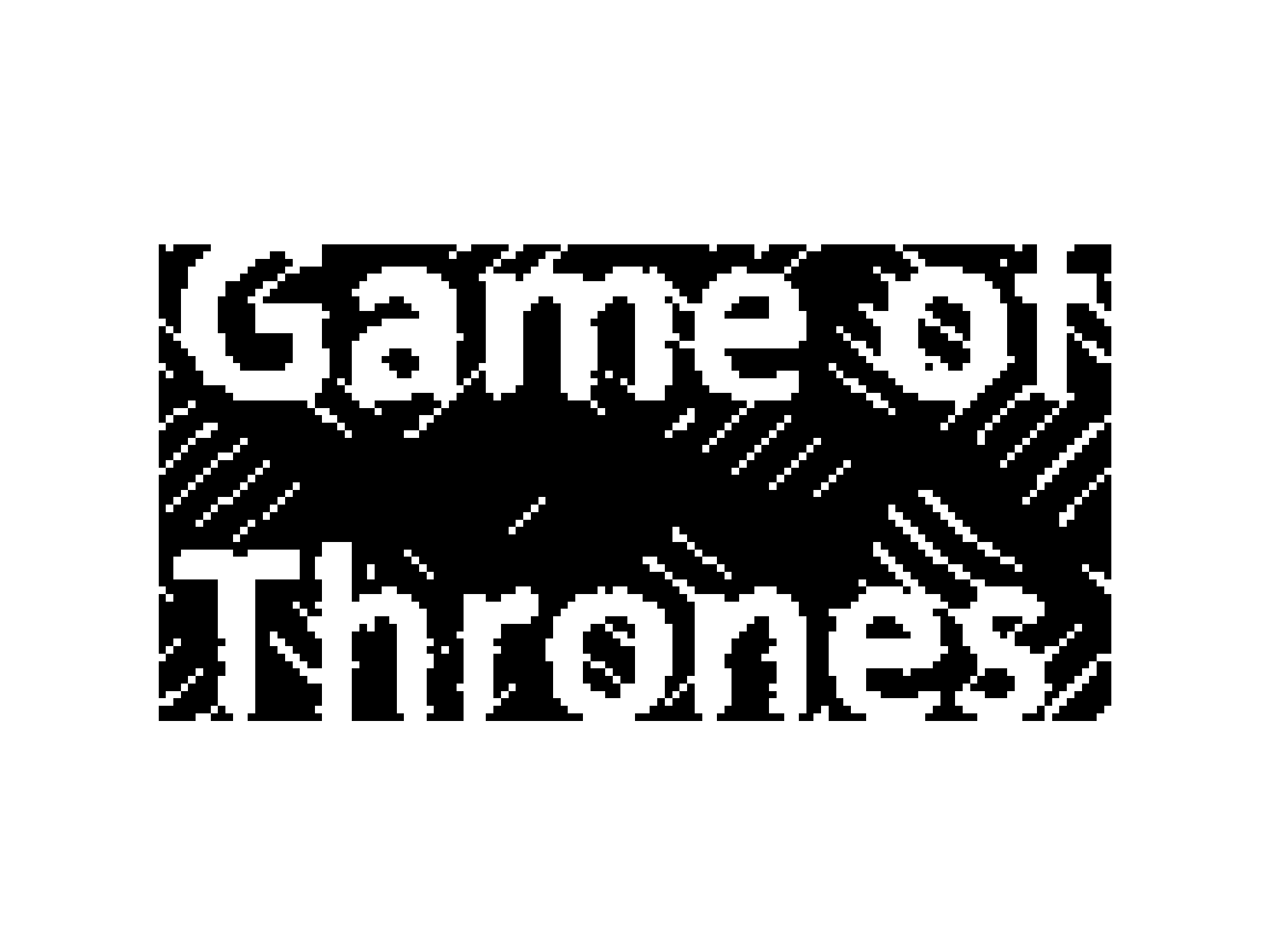}
                 \vspace{-0.61cm}
              \hspace{-4.8cm}
        \end{subfigure}
         \\[1ex]\vspace{-0.9cm}
        \begin{subfigure}[b]{0.22\textwidth}
                \includegraphics[width=\textwidth]{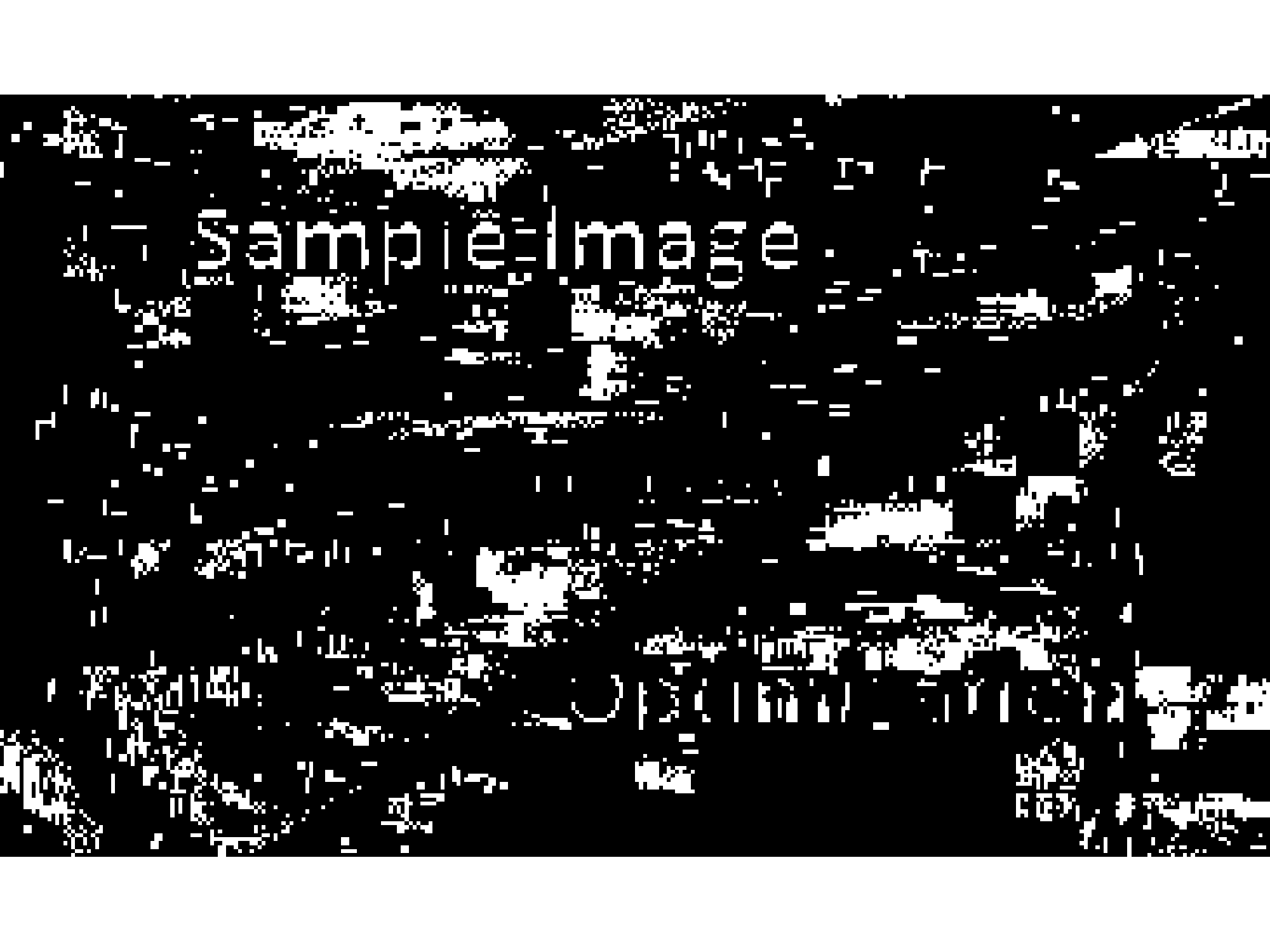}
                                \vspace{-0.32cm}
          \hspace{-1.5cm}    
        \end{subfigure}%
        ~ 
		\vspace{-0.02cm}        
        \begin{subfigure}[b]{0.18\textwidth}
                \includegraphics[width=\textwidth]{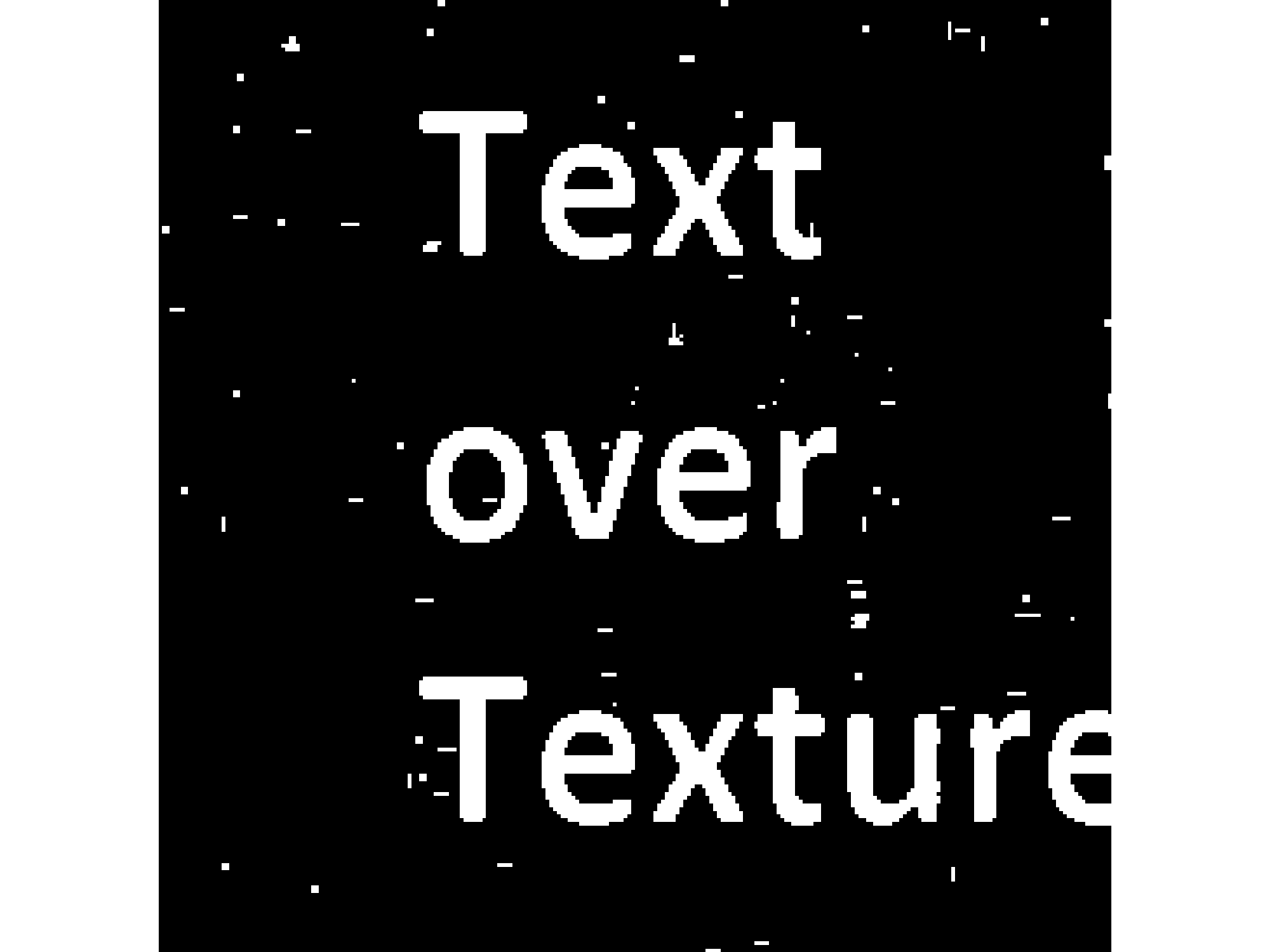}
                \vspace{-0.04cm}
            \hspace{-6cm} 
        \end{subfigure}%
        \begin{subfigure}[b]{0.25\textwidth}
			~ 
            \vspace{-0.42cm}
                \includegraphics[width=\textwidth]{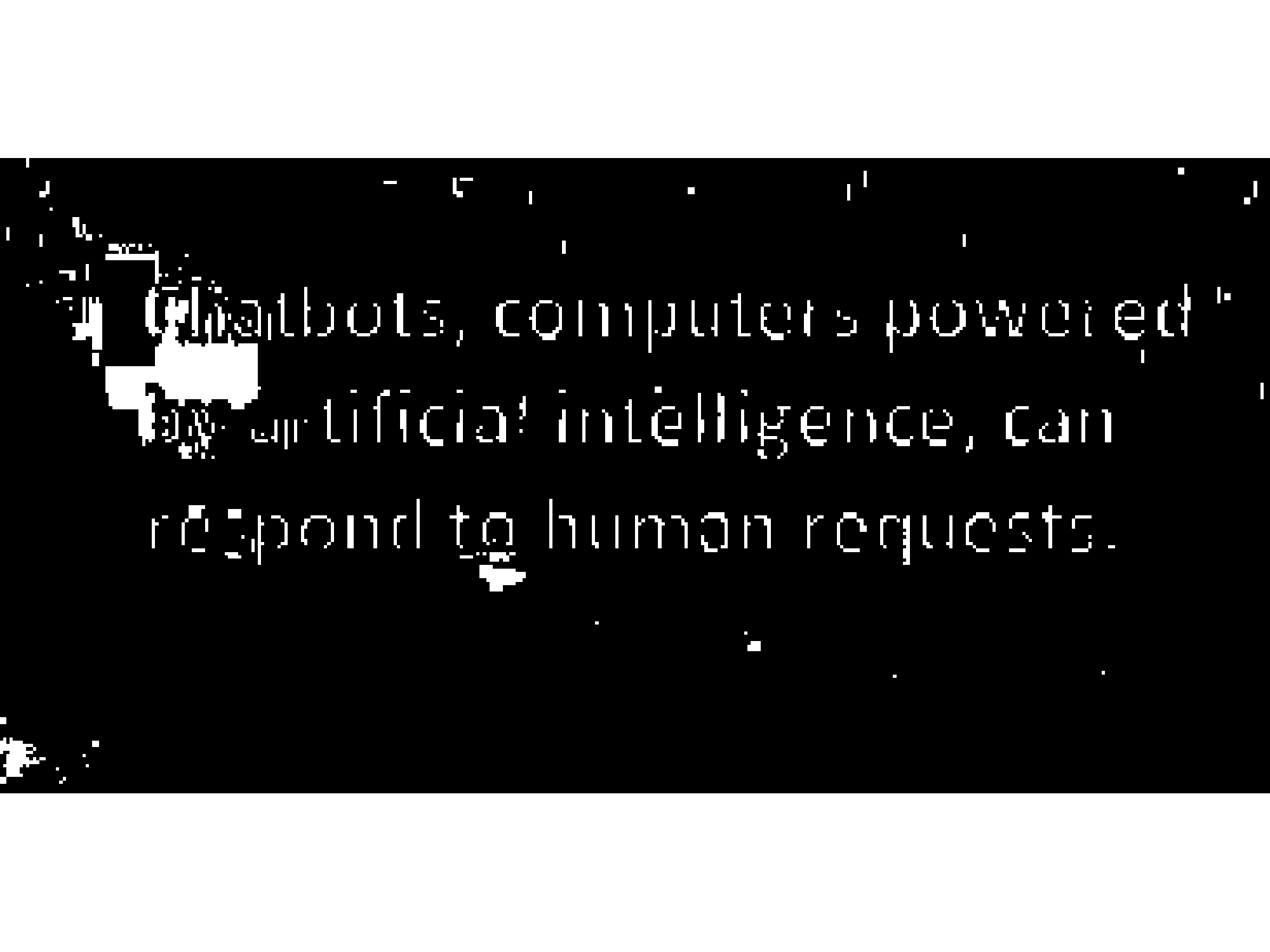}
                \vspace{-0.4cm}
            \hspace{-2cm} 
        \end{subfigure}%
        \begin{subfigure}[b]{0.30\textwidth}
                \includegraphics[width=\textwidth]{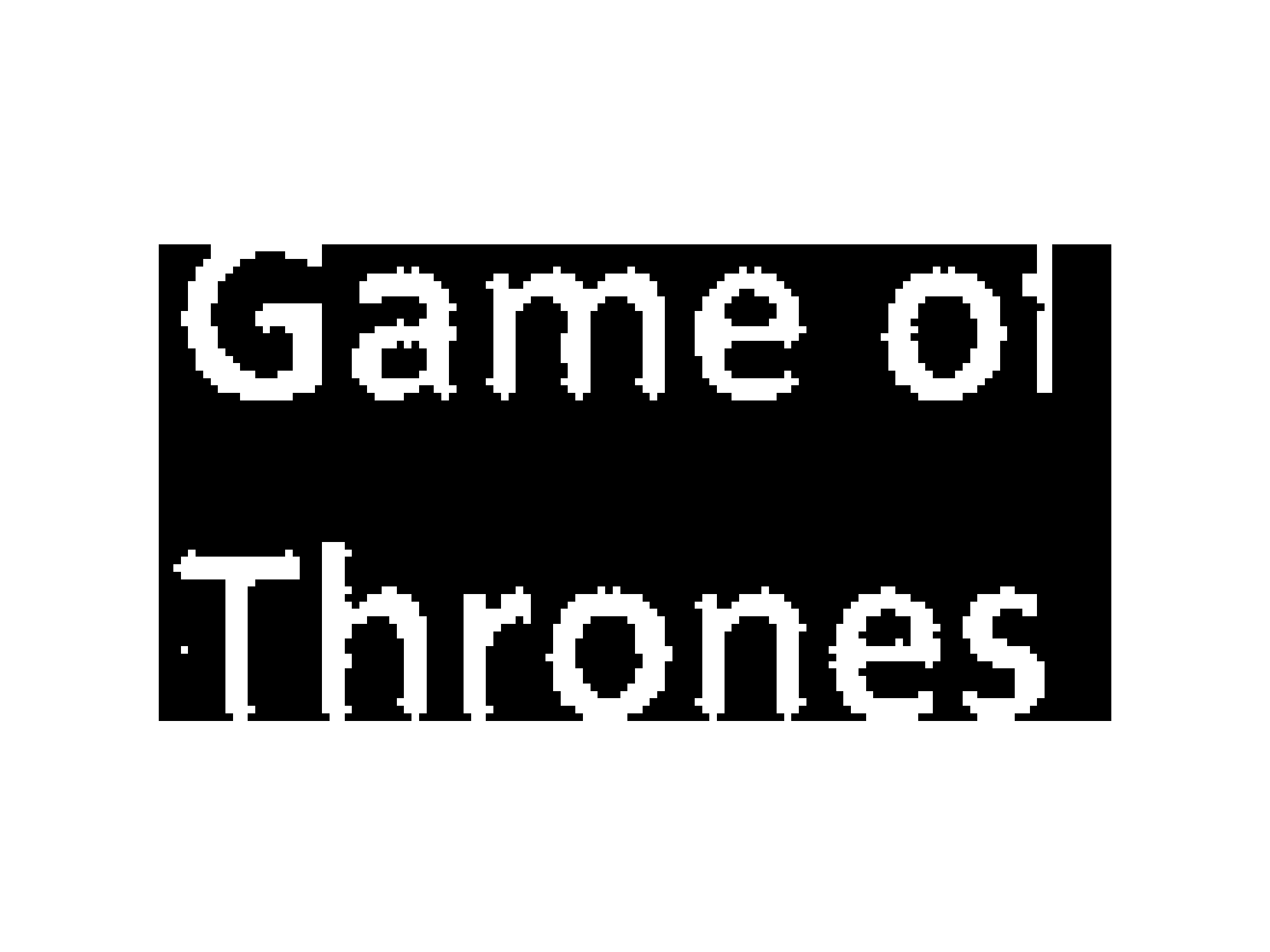}
                 \vspace{-0.61cm}
              \hspace{-4.8cm}
        \end{subfigure}
         \\[1ex]\vspace{-0.9cm}
         \begin{subfigure}[b]{0.22\textwidth}
                \includegraphics[width=\textwidth]{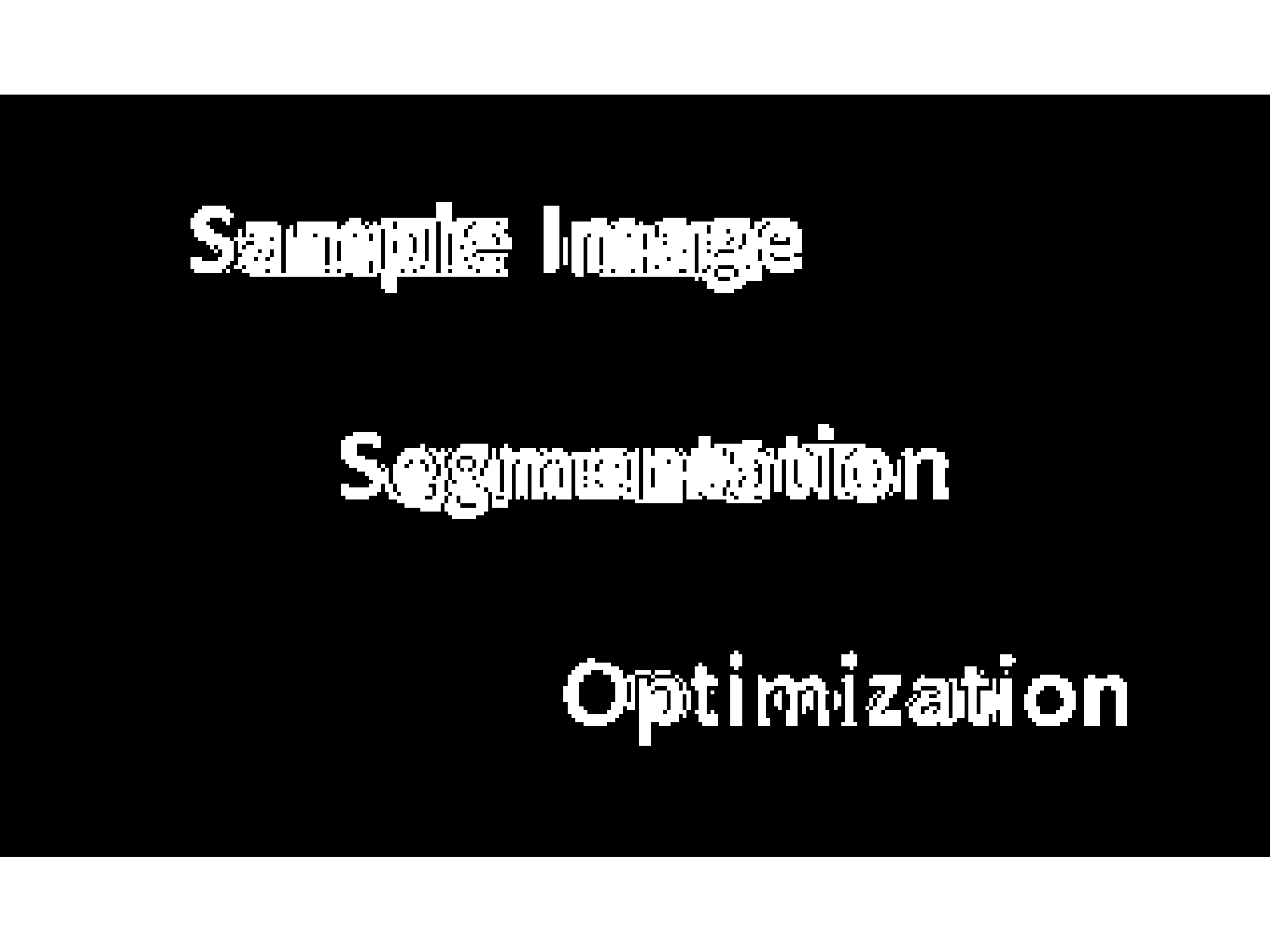}
                                \vspace{-0.32cm}
          \hspace{-1.5cm}    
        \end{subfigure}%
        ~ 
		\vspace{-0.02cm}        
        \begin{subfigure}[b]{0.18\textwidth}
                \includegraphics[width=\textwidth]{texture8_TV-eps-converted-to.pdf}
                \vspace{-0.04cm}
            \hspace{-6cm} 
        \end{subfigure}%
        \begin{subfigure}[b]{0.25\textwidth}
			~ 
            \vspace{-0.42cm}
                \includegraphics[width=\textwidth]{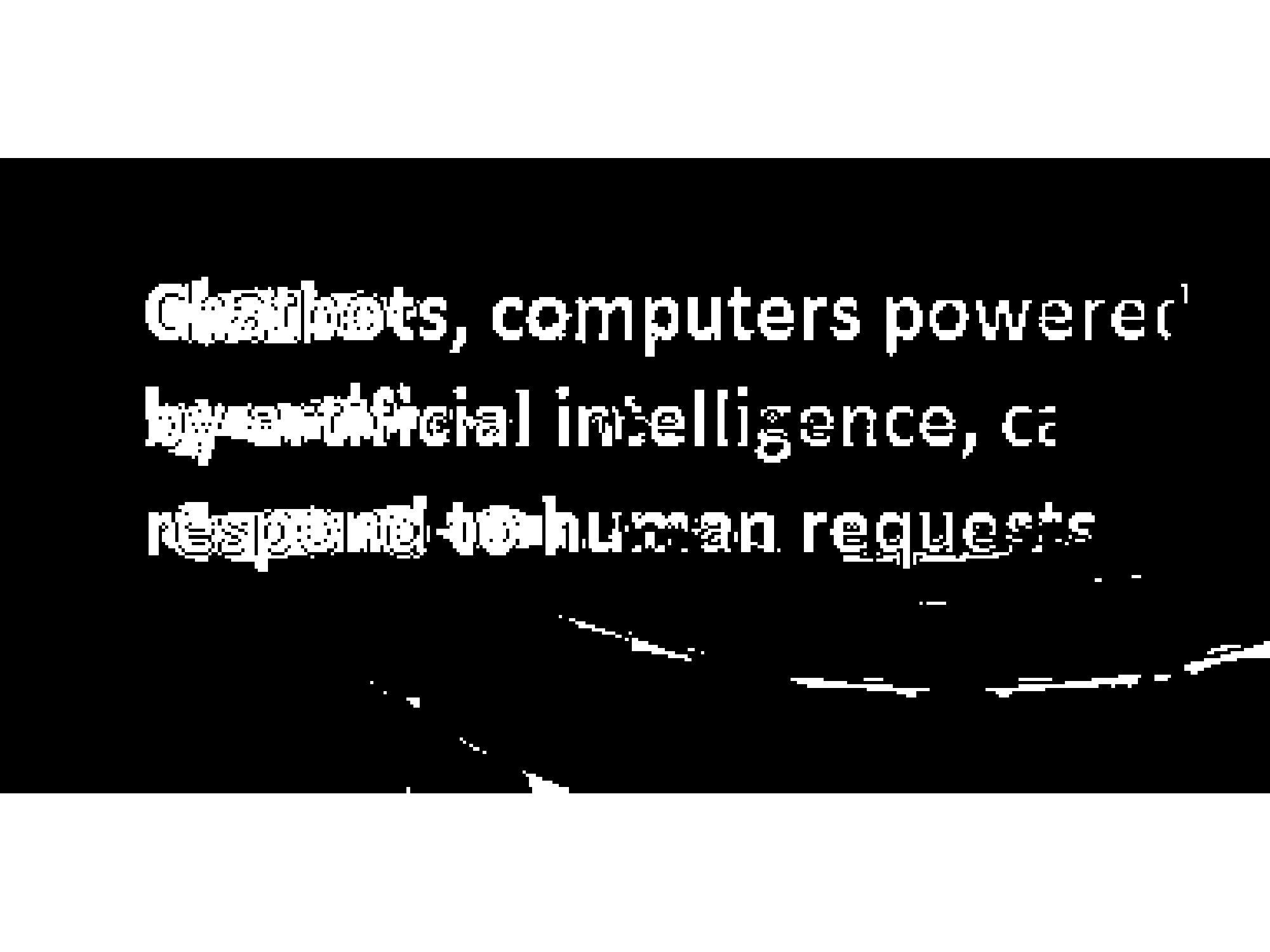}
                \vspace{-0.4cm}
            \hspace{-2cm} 
        \end{subfigure}%
        \begin{subfigure}[b]{0.30\textwidth}
                \includegraphics[width=\textwidth]{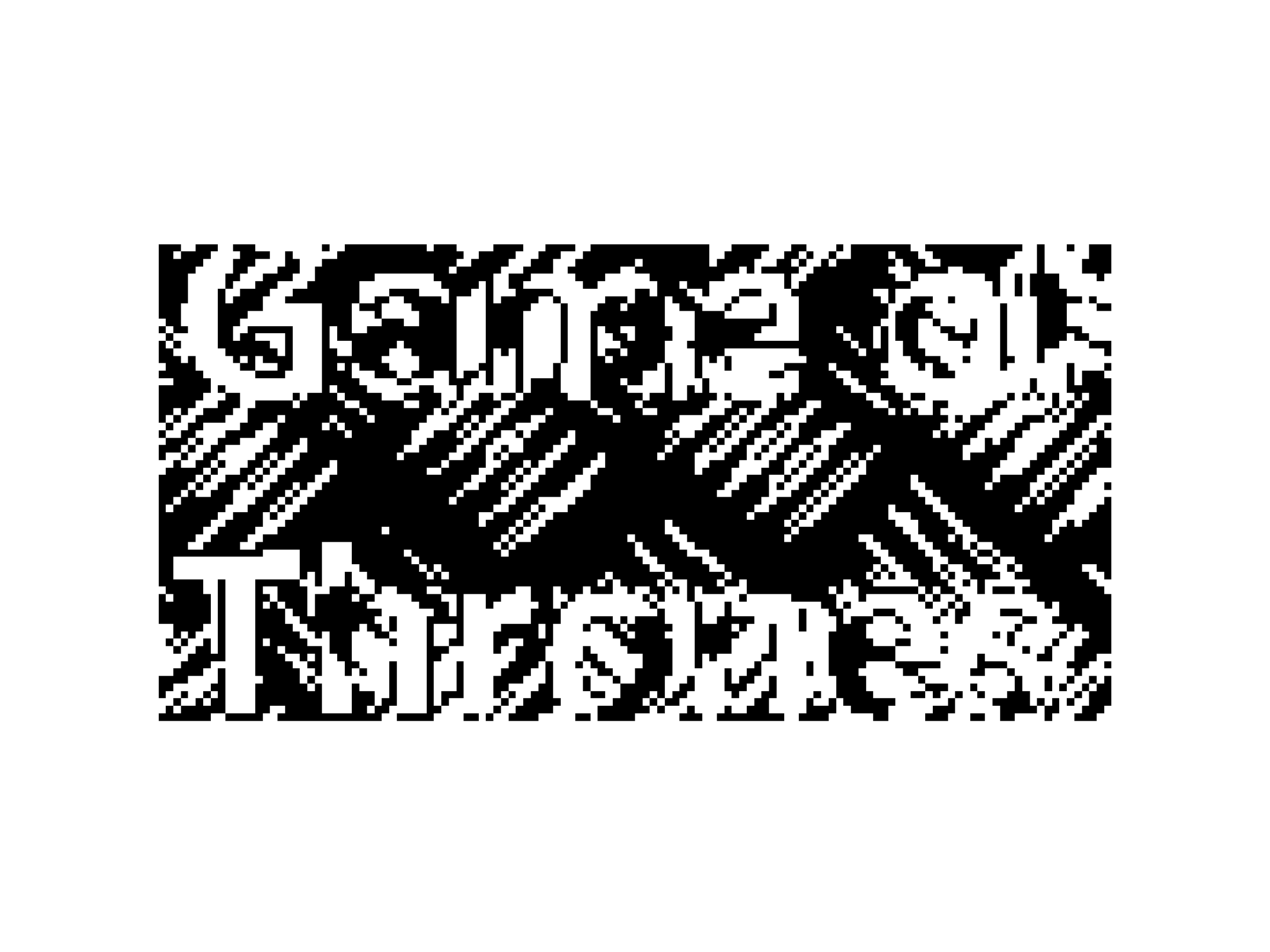}
                 \vspace{-0.61cm}
              \hspace{-4.8cm}
        \end{subfigure}
         \\[1ex]\vspace{-0.9cm}
                 \begin{subfigure}[b]{0.22\textwidth}
                \includegraphics[width=\textwidth]{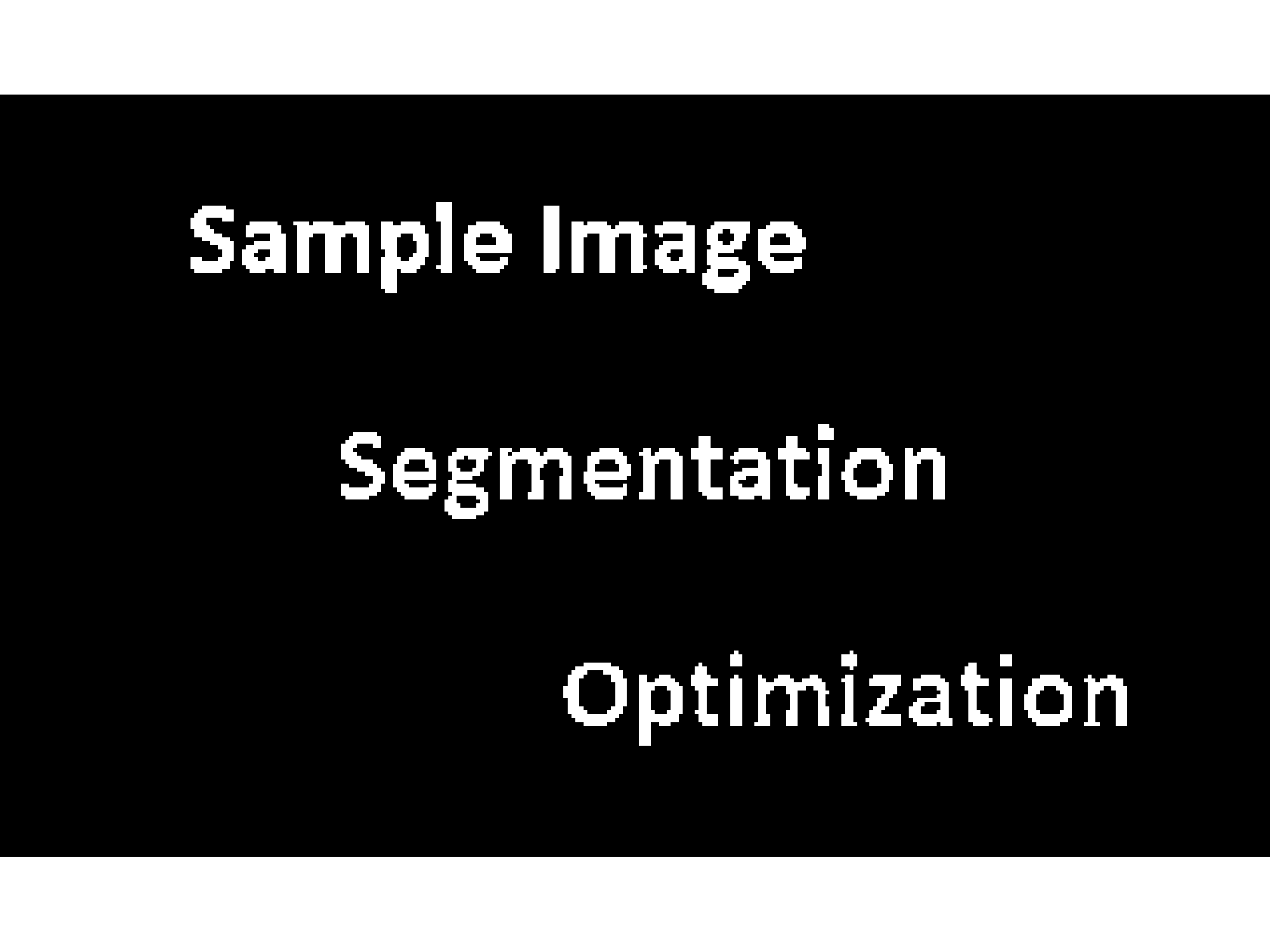}
                                \vspace{-0.3cm}
          \hspace{-1.5cm}    
        \end{subfigure}%
        ~ 
		\vspace{-0.02cm}        
        \begin{subfigure}[b]{0.18\textwidth}
                \includegraphics[width=\textwidth]{mask_texture8-eps-converted-to.pdf}
                \vspace{-0.04cm}
            \hspace{-6cm} 
        \end{subfigure}%
        \begin{subfigure}[b]{0.25\textwidth}
			~ 
            \vspace{-.42cm}
                \includegraphics[width=\textwidth]{mask_test2-eps-converted-to.pdf}
                \vspace{-0.4cm}
            \hspace{-2cm} 
        \end{subfigure}%
        \begin{subfigure}[b]{0.30\textwidth}
                \includegraphics[width=\textwidth]{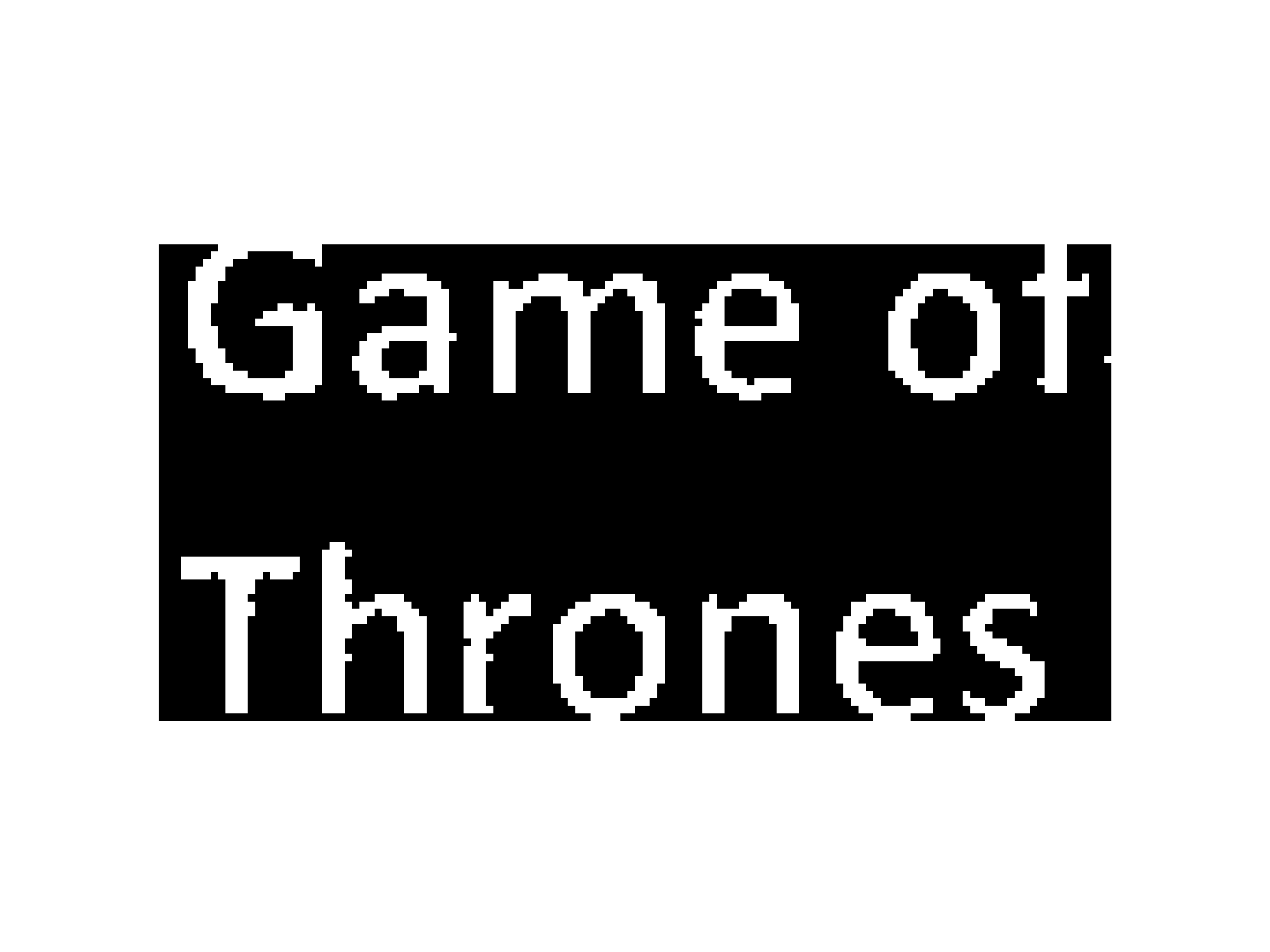}
                 \vspace{-0.61cm}
              \hspace{-4.8cm}
        \end{subfigure}
        \caption{Segmentation result for the selected test images. The images in the first row denotes the original images. And the images in the second, third, forth and the fifth rows denote the foreground map by hierarchical k-means clustering \cite{djvu}, shape primitive extraction and coding \cite{spec}, sparse decomposition \cite{myTV} and the proposed algorithm respectively.}
\end{figure*}

We also provide the average precision, recall and F1 score \cite{metrics} achieved by different algorithms for the above sample images. The average precision, recall and F1 score by different algorithms are given in Table 1.
The precision, recall and F1 score are defined as in Eq. (12) and (13), where TP, FP and FN denote true positive, false positive and false negative respectively. 
In our evaluation, we treat the text pixels as positive. A pixel that is correctly identified as text (compared to the ground-truth) is considered true positive.
\begin{gather}
 \text{Precision}= \frac{\text{TP}}{\text{TP+FP}} \ , 
\ \ \ \ \text{Recall}= \frac{\text{TP}}{\text{TP+FN}} 
\end{gather}
\begin{gather}
\text{F1}= 2 \ \frac{\text{precision} \times \text{recall}}{\text{precision+recall}}
\end{gather}

\begin{table}[ht]
\centering
  \caption{Comparison of accuracy of different algorithms}
  \centering
\begin{tabular}{|m{3.4cm}|m{1.2cm}|m{1.2cm}|m{1.2cm}|}
\hline
Segmentation Algorithm  &  \  \ Precision & \ \  Recall & \  F1 score\\
\hline
 SPEC \cite{spec} & \ \ \ 67\%  & \ \ \ 77\%  & \ \ \  71.6\%\\
\hline
 Hierarchical Clustering \cite{djvu} & \ \ \ 66.5\% & \ \ \ 92\% & \ \ \ 77.2\% \\
\hline 
 Sparse Dec. with TV \cite{myTV} & \ \ \  71\% & \ \ \  91.7\% & \ \ \  80\% \\
\hline
 The proposed algorithm & \ \ \ 95\%  & \ \ \ 92.5\%  & \ \ \  93.7\%\\
\hline
\end{tabular}
\label{TblComp}
\end{table}

As it can be seen, the proposed scheme achieves much higher precision and recall than SPEC, hierarchical k-means clustering and sparse decomposition approach.
It is worth mentioning that our algorithm is pretty robust to the initialized value of variables. 
A complete study of the initialization impact on the segmentation result is presented in \cite{mask_journal}.

\section{conclusion}
This paper proposes a text extraction algorithm from a signal decomposition perspective. We consider texts as an overlaying component on top of a natural scene image, where the pixel values at each point comes from one and only one of the components
(in contrast with the traditional signal decomposition case, where the signal component at each point is assumed to be the summation of corresponding values from different components).
Each component is assumed to have a sparse representation with respect to a suitable subspace.
The text component is also assumed to be sparse.
We then propose an optimization framework to separate the background and text components using the alternating direction method.
Experimental results show that the proposed algorithm can provide significantly better text extraction when the background is textured and has similar color distribution to text.
This algorithm could be further improved by learning the subspaces for the desired application.

\section*{Acknowledgment}
The authors would like to thank Ivan Selesnick, 	
Pablo Sprechmann, and Arian Maleki for their valuable comments regarding this work.

\end{document}